\documentclass{article}

\PassOptionsToPackage{numbers,sort&compress}{natbib}
\usepackage[main, final]{neurips_2026}
\usepackage{makecell}
\usepackage{multirow}
\usepackage{booktabs}
\usepackage{xcolor}
\usepackage{wrapfig}
\usepackage{subcaption}
\usepackage{placeins}
\usepackage[table]{xcolor}
\usepackage{pifont}
\usepackage{xspace}
\definecolor{RedColor}{rgb}{0.949,0.275,0.224}
\definecolor{GreenColor}{rgb}{0.137,0.573,0.565}

\newcommand{\xmark}{\textcolor{RedColor}{\ding{55}}\xspace}
\newcommand{\cmark}{\textcolor{GreenColor}{\ding{51}}\xspace}

\usepackage[utf8]{inputenc}
\usepackage[T1]{fontenc}
\usepackage{enumitem}
\usepackage{hyperref}
\usepackage{url}
\usepackage{booktabs}
\usepackage{amsfonts}
\usepackage{amsmath}
\usepackage{amssymb}
\usepackage{nicefrac}
\usepackage{microtype}
\usepackage{xcolor}
\usepackage{graphicx}
\usepackage{pifont}
\newcommand{\yesmark}{\cmark}
\newcommand{\nomark}{\xmark}
\title{HIST3R: Decomposing and Rectifying State Updates with History for Streaming 3D Reconstruction}

\author{%
  Kenneth Yang$^{1}$ \qquad Yu-Chiang Frank Wang$^{1,2}$ \\
  $^{1}$National Taiwan University \qquad $^{2}$NVIDIA \\
  \texttt{r13942093@ntu.edu.tw} \qquad \texttt{frankwang@nvidia.com}
}
\begin{document}

\maketitle

\begin{abstract}
Streaming 3D reconstruction with recurrent states enables long-video scene understanding under bounded memory, but turns reconstruction into a state-evolution problem: each incoming frame updates a compact scene memory on which all subsequent predictions depend, so unreliable updates overwrite historical context and accumulate into long-horizon geometric drift. We analyze CUT3R-style recurrent updates through a direction--magnitude decomposition, in which the magnitude sets the update gain, i.e., how strongly an observation is written into the state, and the direction encodes the geometric intent, i.e., where the update drives the state. This decomposition provides a unified view of CUT3R and its variants, revealing that existing training-free extensions perform magnitude-only control: they can attenuate an unreliable update but cannot correct its direction. We propose HIST3R, a training-free framework built on CUT3R that controls both components using history. HIST3R rectifies the update direction using locally consistent historical states, and modulates the update magnitude according to the visual discrepancy between the current observation and its history, suppressing redundant updates while preserving complementary evidence. Across camera pose estimation, 3D reconstruction, and video depth estimation, HIST3R consistently improves long-horizon performance while retaining the memory footprint and inference speed of CUT3R. With 1000 input frames, it reduces ATE by 23.87\% on ScanNet and 33.65\% on TUM-Dynamics relative to the strongest competing method; with 500 frames, it improves median reconstruction accuracy over the strongest CUT3R-based variant by 56.52\% on 7-Scenes and 33.96\% on NRGBD. 

\textit{Project page:\url{https://kennethyangg.github.io/HIST3R}.} 

\end{abstract}

\section{Introduction}
Recovering dense 3D geometry from 2D observations is a central problem in computer vision, supporting applications across autonomous driving~\citep{wei2023surroundocc,huang2023tpvformer,tian2023occ3d}, augmented reality (AR)~\citep{newcombe2011kinectfusion,hong2024clothed,zheng2024gpsgaussian,lei2025mosca}, and embodied AI~\citep{wu2025embodiedocc,han2022scene,yan2023active}. Recent Transformer-based foundation models~\citep{wang2024dust3r,leroy2024mast3r,wang2025vggt} have increasingly replaced classical optimization-based pipelines~\citep{snavely2006photo,schonberger2016sfm,seitz2006mvs,furukawa2010pmvs,schonberger2016mvs} due to their strong generalization ability and efficient feed-forward inference. Yet this advantage does not scale to long videos: reasoning jointly over an ever-growing set of views makes feed-forward reconstruction increasingly memory- and compute-intensive.

Streaming 3D reconstruction addresses this scalability issue by processing frames sequentially~\citep{wu2025point3r,streamvggt2026,stream3r2026,yuan2026infinitevggt,wang2025cut3r,chen2026ttt3r,zheng2026ttsa3r,li2026wint3r,xu2026framevggt}. Existing methods preserve historical context through either \textbf{explicit memory} or \textbf{implicit memory}. Explicit-memory methods retain processed frames, features, or tokens in an external memory bank, preserving rich history at the cost of memory or computation that grows with sequence length~\citep{wu2025point3r,streamvggt2026,stream3r2026}. Implicit-memory methods, including ours, instead compress history into a fixed-size recurrent state with bounded memory~\citep{wang2025cut3r,chen2026ttt3r,zheng2026ttsa3r}. This compression, however, shifts the central challenge from per-frame reconstruction to recurrent state evolution: each incoming frame updates the only compact memory of the scene, so an unreliable update can overwrite useful history, and over many frames such errors accumulate into long-horizon geometric drift~\citep{chen2026ttt3r,zheng2026ttsa3r}.

We study this update problem through CUT3R~\citep{wang2025cut3r}, a representative implicit-memory streaming model. For each incoming frame, CUT3R produces a candidate update that writes the new observation into the recurrent state. Such an update must determine two things: how strongly the observation should modify the accumulated scene representation, and what geometric content it should contribute. We refer to the former as the \textbf{update gain} and the latter as the \textbf{geometric intent}. In the recurrent state space, these two roles correspond naturally to the magnitude and direction of the candidate update, which motivates a \emph{direction--magnitude decomposition} of recurrent state updates.

\begin{figure*}[!t]
    \centering
    \includegraphics[
        width=1.0\textwidth,
    ]{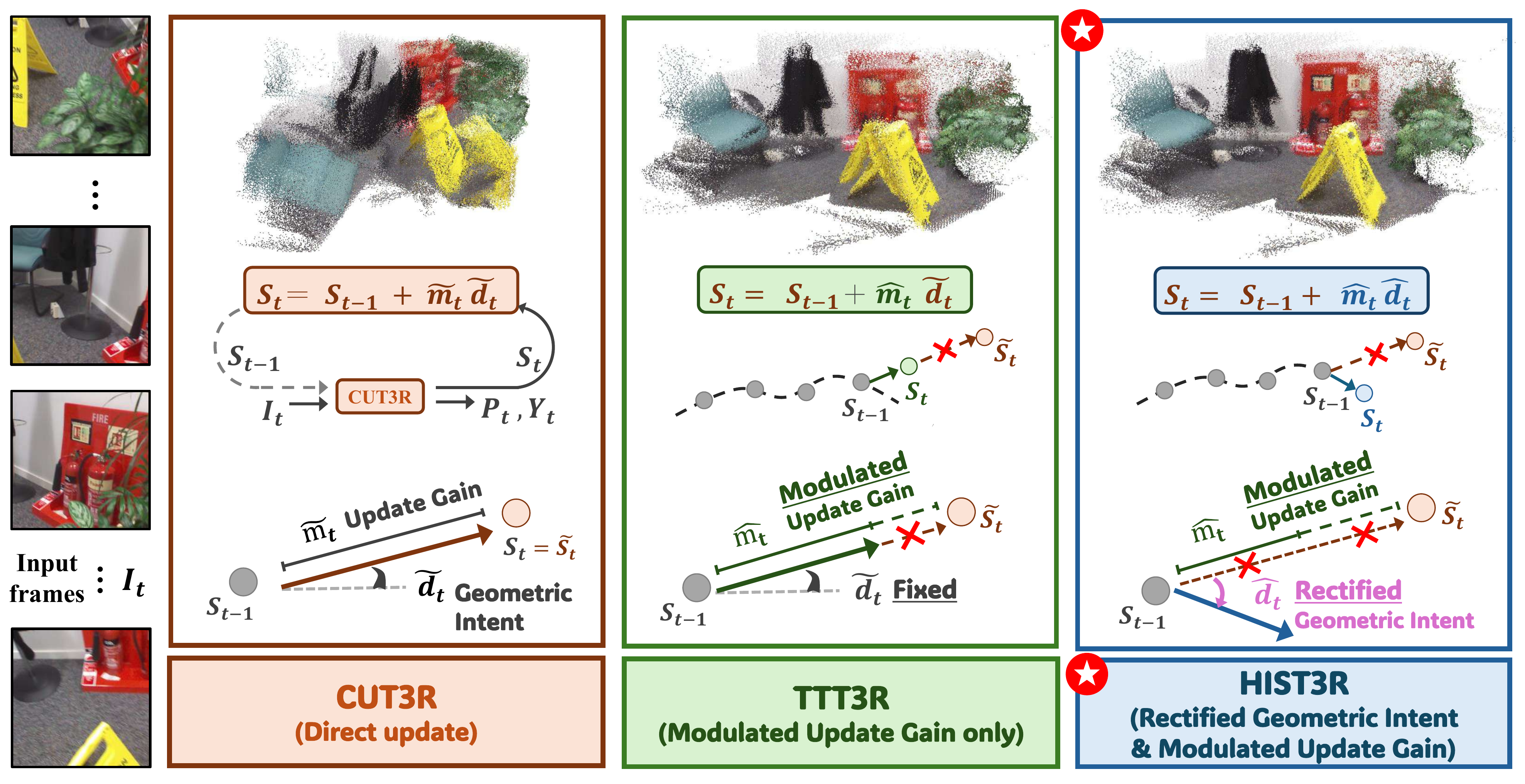}
    \caption{
    \textbf{A unified direction--magnitude view of recurrent state updates for long-sequence streaming 3D reconstruction.}
    Update magnitude and direction correspond to update gain and geometric intent, respectively. CUT3R~\citep{wang2025cut3r} directly commits the candidate state, while magnitude-only CUT3R variants, represented by TTT3R~\citep{chen2026ttt3r}, rescale the candidate update magnitude but keep its direction fixed. HIST3R controls both components by rectifying the update direction and modulating the update magnitude, enabling more reliable long-horizon state evolution.
    }
    \label{fig:teaser}
\end{figure*}
As illustrated in Fig.~\ref{fig:teaser}, this decomposition provides a unified view of CUT3R and its recent training-free extensions. Vanilla CUT3R commits the candidate update as is, while TTT3R~\citep{chen2026ttt3r} and TTSA3R~\citep{zheng2026ttsa3r} can be interpreted as magnitude-only update control: they estimate a scalar gate from auxiliary signals, such as state-to-image attention or temporal--spatial dissimilarity, to modulate the update gain while leaving the update direction unchanged. As a result, they can attenuate an unreliable or redundant update, but cannot correct the geometric intent that determines where the recurrent state is driven. This limitation becomes increasingly critical over long input sequences. When an incoming frame induces a biased candidate update, magnitude-only gating may reduce its step size, yet the state still moves along the same biased direction. Since the recurrent state compactly carries past information, such directional biases persist and compound across frames, gradually pulling the state away from coherent scene structure and leading to geometric drift. Reliable long-horizon streaming reconstruction can therefore be viewed as a state-space evolution problem, requiring control over both how strongly the recurrent state is updated and how its geometric evolution is directed.

Motivated by this view, we propose \textbf{HIST3R}, a training-free framework built on CUT3R for long-sequence streaming 3D reconstruction. HIST3R decomposes the CUT3R candidate update into \emph{direction} and \emph{magnitude}, enabling separate control over geometric intent and update gain. To rectify geometric intent, HIST3R uses recent recurrent states as a local reference for plausible state evolution and adjusts the observation-induced update toward a history-consistent endpoint. To modulate the update gain, HIST3R estimates how much new evidence the incoming observation provides to the existing recurrent state. It therefore compares the current visual context with historical visual contexts and adapts the update magnitude according to their discrepancy, suppressing redundant state changes while preserving responsiveness to observations that depart from recent history.

We evaluate HIST3R across camera pose estimation, 3D reconstruction, and video depth estimation under both standard and long-sequence streaming settings. Across these evaluations, HIST3R consistently improves long-horizon performance over prior streaming methods while maintaining competitive short-sequence accuracy, all without additional training. These results demonstrate that direction--magnitude decomposition provides an effective mechanism for controlling geometric intent and update gain in implicit-memory streaming reconstruction.

Our contributions are summarized as follows:
\begin{itemize}
\item We introduce a direction--magnitude decomposition of recurrent state updates, which unifies CUT3R and its variants and reveals update direction as the missing control axis for long-horizon reliability.

\item We propose HIST3R, a training-free extension of CUT3R that rectifies geometric intent with historical recurrent states and modulates update gain with observation--history visual discrepancy, while preserving the memory footprint and inference speed of CUT3R.

\item We demonstrate consistent long-horizon improvements across camera pose estimation, 3D reconstruction, and video depth estimation. At 1000 input frames, HIST3R reduces ATE over the second-best method, TTSA3R, by \(\mathbf{23.87\%}\) on ScanNet and \(\mathbf{33.65\%}\) on TUM-Dynamics; at 500 frames, it improves median reconstruction accuracy over TTSA3R by \(\mathbf{56.52\%}\) on 7-Scenes and \(\mathbf{33.96\%}\) on NRGBD.

\end{itemize}

\section{Related Work}
\paragraph{Offline Reconstruction Foundation Models.}
Recent learning-based methods have shifted 3D reconstruction from classical geometric pipelines~\citep{snavely2006photo,schonberger2016sfm,seitz2006mvs,furukawa2010pmvs,schonberger2016mvs} toward generalizable feed-forward foundation models~\citep{wang2024dust3r,leroy2024mast3r,zhang2024monst3r,yang2025fast3r,tang2025mvdust3r,wang2025vggt,zhang2025flare}. One line of work builds on pairwise reconstruction. DUSt3R~\citep{wang2024dust3r} formulates 3D reconstruction as direct pointmap regression from image pairs, and subsequent methods improve matching, dynamic-scene robustness, and reconstruction quality within this pairwise paradigm~\citep{leroy2024mast3r,zhang2024monst3r,chen2025easi3r}. Although effective for local two-view geometry, these methods still require global alignment or aggregation to reconstruct larger image collections.

Another line of work performs all-at-once multi-view reconstruction. VGGT~\citep{wang2025vggt} jointly processes multiple images with global transformer attention to predict camera poses, depth, pointmaps, and tracks, with recent variants improving inference speed, geometric priors, or scalability~\citep{shen2026fastvggt,wang2026pi3,lin2026depthanything3,li2026svggt,wang2026flashvggt}. However, offline and all-at-once models assume a fixed input set and reason over all views jointly, making memory and computation grow rapidly as more frames are added. This scalability bottleneck motivates online reconstruction foundation models, which process frames incrementally while maintaining historical scene information.

\paragraph{Online Reconstruction Foundation Models.}
Online reconstruction foundation models can be broadly grouped into explicit-memory and implicit-memory approaches.
Explicit-memory methods retain processed frames as externally accessible entries, such as spatial memories, cached features, or historical tokens.~\citep{wang2025spann3r,wu2025point3r,streamvggt2026,stream3r2026,yuan2026infinitevggt,li2026wint3r,xu2026framevggt}. 
Spann3R~\citep{wang2025spann3r} adapts DUSt3R into an online setting by storing previous predictions as key-value features, while Point3R~\citep{wu2025point3r} introduces a spatial pointer memory associated with 3D positions. 
StreamVGGT~\citep{streamvggt2026} and STream3R~\citep{stream3r2026} instead use causal attention with key-value caching to process incoming frames incrementally. 
Although these methods avoid full reprocessing of all frames, their memory or computation still tends to grow with sequence length. 
InfiniteVGGT~\citep{yuan2026infinitevggt} mitigates this issue through token pruning inspired by streaming language models~\citep{xiao2024streamingllm,zhang2023h2o}, but its pruning strategy mainly addresses cache efficiency rather than explicitly correcting accumulated geometric errors.

Implicit-memory methods, represented by CUT3R~\citep{wang2025cut3r}, compress history into a fixed-size recurrent state, keeping memory bounded regardless of sequence length~\citep{wang2025cut3r,chen2026ttt3r,zheng2026ttsa3r} . Since all subsequent predictions depend on this compact state, reconstruction quality hinges on how each observation updates it. Recent training-free extensions improve update stability by rescaling the candidate update: TTT3R~\citep{chen2026ttt3r} views the recurrent update as test-time training and derives per-token learning rates from state-to-image attention, while TTSA3R~\citep{zheng2026ttsa3r} derives its gates from temporal--spatial dissimilarity. Both regulate how strongly the candidate update is applied but leave its direction unchanged, corresponding to magnitude-only control under our decomposition. In contrast, HIST3R uses history to control both the direction and the magnitude of the update, while remaining training-free and preserving the bounded memory of implicit-memory methods.

\section{Method}
\begin{figure*}[t]
    \centering
    \includegraphics[
        width=1.0\textwidth,
        trim=10 5 10 5,
        clip
    ]{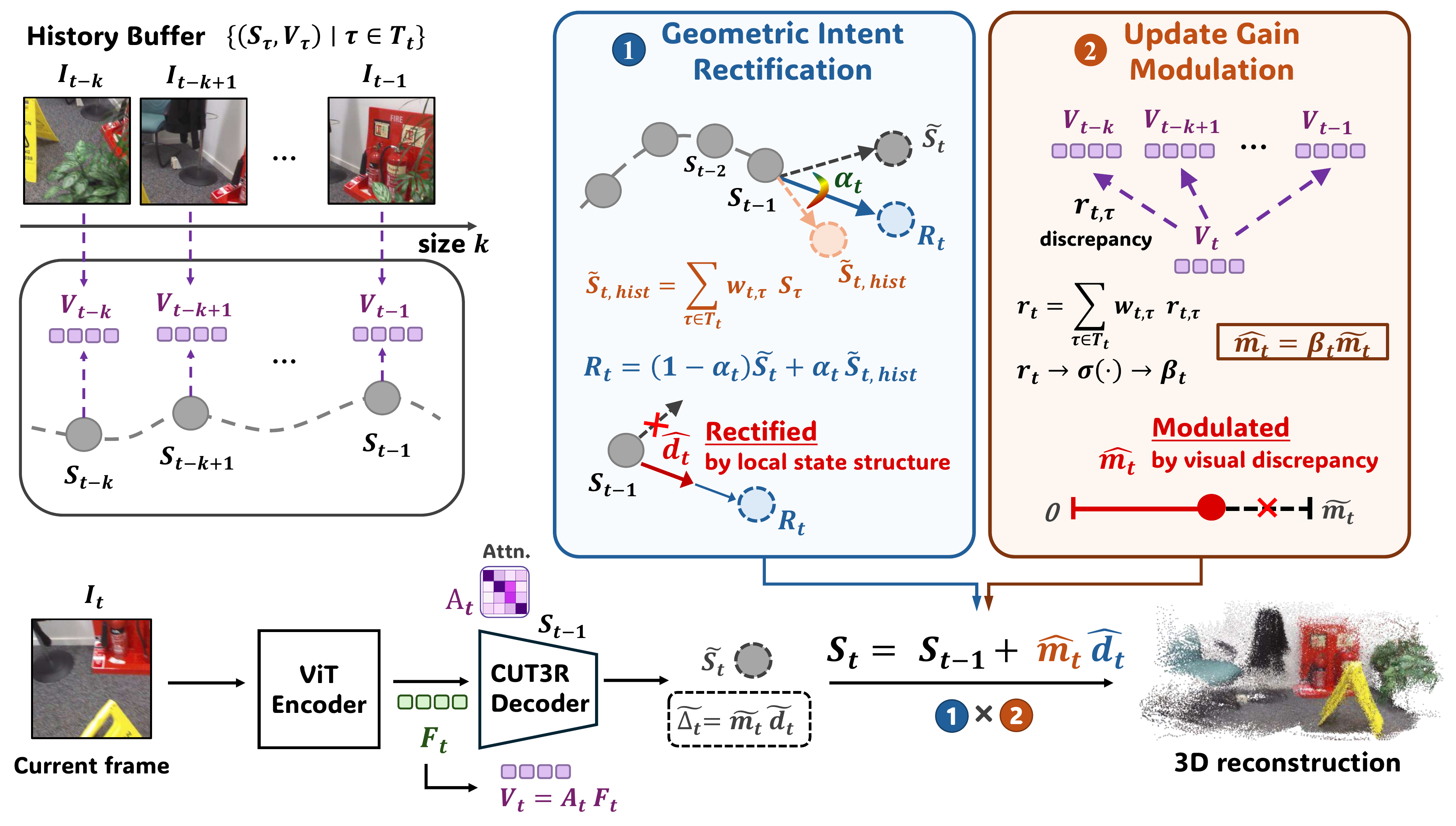}
    \caption{
    \textbf{Overview of HIST3R.} HIST3R maintains a sliding history buffer of past recurrent states and visual summaries. It decomposes the CUT3R candidate update into direction and magnitude to control geometric intent and update gain, respectively. Historical states rectify the update direction, while observation--history visual discrepancy modulates the update magnitude. The two components are combined into the final recurrent state update.
    }
    \label{fig:method_overview}
\end{figure*}

We present \textbf{HIST3R}, a training-free extension of CUT3R for long-sequence streaming 3D reconstruction. As illustrated in Fig.~\ref{fig:method_overview}, HIST3R improves recurrent state updating by explicitly controlling the two factors introduced in Sec.~\ref{sec:decomposition}: geometric intent and update gain.

\subsection{Long-Horizon State Evolution and Update Decomposition}
\label{sec:decomposition}
\paragraph{Recurrent state update in CUT3R.}
Given an online image stream \(\{I_t\}_{t=1}^{T}\), streaming 3D reconstruction predicts the camera pose \(P_t\) and dense pointmap \(Y_t \in \mathbb{R}^{H \times W \times 3}\) for each frame on the fly. 
CUT3R processes the stream with a persistent recurrent state \(S_t \in \mathbb{R}^{N \times C}\), which compactly stores scene history using \(N\) state tokens. 
For each incoming frame, the image tokens \(F_t \in \mathbb{R}^{P \times C}\) interact with the previous state \(S_{t-1}\) through interconnected transformer decoders:
\begin{equation}
    F'_t, \tilde{S}_t
    =
    \mathrm{Decoders}(F_t, S_{t-1}).
    \label{eq:state_image_interaction}
\end{equation}
The decoders perform a dual interaction: image tokens read historical information from \(S_{t-1}\) to produce history-conditioned tokens \(F'_t\), while state tokens integrate visual evidence from \(F_t\) to form the candidate state \(\tilde{S}_t\). 
The resulting \(F'_t\) is fed into task-specific heads to predict \(P_t\) and \(Y_t\).

Vanilla CUT3R commits the candidate state directly, \(S_t=\tilde{S}_t\), which applies the candidate update \(\tilde{\Delta}_t=\tilde{S}_t-S_{t-1}\) to the previous state without revision. 
Since \(S_t\) conditions all subsequent predictions, an unreliable candidate update, caused by factors such as motion blur, occlusion, or challenging viewpoints, is written directly into the compact scene memory and can weaken the historical context it encodes~\cite{chen2026ttt3r, zheng2026ttsa3r}.

\paragraph{Direction--magnitude decomposition.}
Each recurrent update plays two roles in reconstruction: it determines what geometric content the current observation contributes to the accumulated scene representation, which we call the \emph{geometric intent}, and how strongly this content is committed, which we call the \emph{update gain}. 
In recurrent state space, these two roles are naturally carried by the \emph{direction} and \emph{magnitude} of the candidate update, respectively. 
We therefore decompose the candidate update of each state token \(i\) as
\begin{equation}
\begin{aligned}
    \tilde{\Delta}_t^{(i)}
    &= \tilde{S}_t^{(i)} - S_{t-1}^{(i)}
     = \tilde{m}_t^{(i)} \tilde{d}_t^{(i)}, \\
    \tilde{d}_t^{(i)}
    &=
    \frac{\tilde{\Delta}_t^{(i)}}{\tilde{m}_t^{(i)}+\epsilon}, \qquad
    \tilde{m}_t^{(i)}
    = \|\tilde{\Delta}_t^{(i)}\|_2,
\end{aligned}
\label{eq:decomposition}
\end{equation}
where \(\tilde{d}_t^{(i)}\) and \(\tilde{m}_t^{(i)}\) denote the candidate geometric intent and candidate update gain, respectively.

This decomposition provides a unified update-control view of CUT3R and its variants. 
Vanilla CUT3R commits both the candidate geometric intent and update gain without revision. 
TTT3R and TTSA3R~\cite{chen2026ttt3r,zheng2026ttsa3r} can be interpreted as \emph{magnitude-only} modulation: they estimate a scalar gate \(g_t^{(i)} \in [0,1]\) from auxiliary signals, such as attention-based confidence or temporal--spatial dissimilarity, and use it to modulate the update gain while keeping the geometric intent fixed:
\begin{equation}
    S_t^{(i)}
    =
    S_{t-1}^{(i)}
    +
    g_t^{(i)} \tilde{m}_t^{(i)} \tilde{d}_t^{(i)}.
    \label{eq:gated_update}
\end{equation}
Such gating can attenuate an unreliable update, but cannot correct where the state is driven: a biased geometric intent continues to drive the state, only with a smaller step. 
Over long sequences, such uncorrected geometric intent accumulates and gradually moves the recurrent state away from coherent evolution.

HIST3R addresses this missing control axis by controlling both components with history. 
It rectifies the geometric intent through historical state consistency to obtain \(\hat{d}_t^{(i)}\), and modulates the update gain through observation--history visual discrepancy to obtain \(\hat{m}_t^{(i)}\). 
The final recurrent update is
\begin{equation}
    S_t^{(i)}
    =
    S_{t-1}^{(i)}
    +
    \hat{m}_t^{(i)} \hat{d}_t^{(i)}.
    \label{eq:hist3r_update}
\end{equation}
Sec.~\ref{sec:direction} and Sec.~\ref{sec:magnitude} describe how \(\hat{d}_t^{(i)}\) and \(\hat{m}_t^{(i)}\) are computed, respectively.
\subsection{Rectifying Geometric Intent via Historical State Consistency}
\label{sec:direction}

The decomposition in Sec.~\ref{sec:decomposition} identifies the geometric intent as the component that determines where the recurrent state evolves. 
A reliable geometric intent should therefore not be determined solely by the current frame, but should also respect how the scene representation has evolved in recent history. 
As shown in Appendix~\ref{app:state_structure}, CUT3R states support using recent history as such a local reference: adjacent states have high cosine affinity, and their variations are largely captured by a low-dimensional subspace. 
Under this local structure, a compatible candidate endpoint should be explainable by nearby historical states, following the self-expressiveness principle in subspace modeling~\citep{elhamifar2013sparse} and classical geometric priors based on local low-dimensional subspaces~\citep{zhu2014complex,kumar2017spatio,kumar2018scalable}.

Based on this observation, HIST3R rectifies the geometric intent by correcting the endpoint that defines it. For each state token \(i\), HIST3R blends the CUT3R candidate endpoint \(\tilde{S}_t^{(i)}\) with a history-proposed endpoint \(\tilde{S}_{t,\mathrm{hist}}^{(i)}\), and derives the rectified direction from the resulting endpoint \(R_t^{(i)}\):
\begin{equation}
    R_t^{(i)}
    =
    (1-\alpha_t^{(i)})\tilde{S}_t^{(i)}
    +
    \alpha_t^{(i)} \tilde{S}_{t,\mathrm{hist}}^{(i)},
    \qquad
    \hat{d}_t^{(i)}
    =
    \frac{
    R_t^{(i)} - S_{t-1}^{(i)}
    }{
    \left\|R_t^{(i)} - S_{t-1}^{(i)}\right\|_2 + \epsilon
    },
    \label{eq:direction_rectified_endpoint}
\end{equation}
where \(\alpha_t^{(i)} \in [0,1]\) is an alignment gate that determines how much history should guide the update with \(\alpha_t^{(i)}=0\) recovering the candidate geometric intent of CUT3R. We next describe how \(\tilde{S}_{t,\mathrm{hist}}^{(i)}\) and \(\alpha_t^{(i)}\) are constructed.

\paragraph{History-proposed endpoint.}
HIST3R maintains a sliding history buffer of the most recent \(k\) recurrent states and visual summaries:
\begin{equation}
    \mathcal{H}_t
    =
    \{(S_\tau, V_\tau) \mid \tau \in \mathcal{T}_t\},
    \qquad
    \mathcal{T}_t
    =
    \{\max(1,t-k), \ldots, t-1\}.
    \label{eq:history_window}
\end{equation}
The cached states \(\{S_\tau\}\) are used here, while the visual summaries \(\{V_\tau\}\) are used in Sec.~\ref{sec:magnitude} to modulate the update gain. 
To construct the history-proposed endpoint, HIST3R retrieves locally similar past states in a token-wise manner. 
It computes cosine affinities \(a_{t,\tau}^{(i)}=\cos(\tilde{S}_t^{(i)}, S_\tau^{(i)})\) for \(\tau \in \mathcal{T}_t\), normalizes them into local history weights, and aggregates the historical states accordingly:
\begin{equation}
    w_{t,\tau}^{(i)}
    =
    \frac{\exp(a_{t,\tau}^{(i)})}
    {\sum_{\tau' \in \mathcal{T}_t}\exp(a_{t,\tau'}^{(i)})},
    \qquad
    \tilde{S}_{t,\mathrm{hist}}^{(i)}
    =
    \sum_{\tau \in \mathcal{T}_t}
    w_{t,\tau}^{(i)} S_\tau^{(i)} .
    \label{eq:history_weights_proposal}
\end{equation}
The resulting \(\tilde{S}_{t,\mathrm{hist}}^{(i)}\) expresses the candidate endpoint through nearby historical states, reflecting the structure of recent state evolution.

\paragraph{Alignment-aware rectification.}
The history-proposed endpoint should guide the update only when it agrees with the current candidate. 
High alignment indicates that recent states provide a reliable reference for rectifying the candidate geometric intent, while low alignment suggests that the current observation introduces a genuine change and should not be over-constrained by history. 
We therefore define the alignment gate as
\begin{equation}
    c_t^{(i)}
    =
    \cos\!\left(\tilde{S}_t^{(i)}, \tilde{S}_{t,\mathrm{hist}}^{(i)}\right),
    \qquad
    \alpha_t^{(i)}
    =
    \sigma\!\left(
    \lambda_{\mathrm{align}}
    \left(c_t^{(i)}-\theta_{\mathrm{align}}\right)
    \right),
    \label{eq:direction_gate}
\end{equation}
where \(\lambda_{\mathrm{align}}\) and \(\theta_{\mathrm{align}}\) control the sharpness and threshold of the gate. 
The resulting \(\hat{d}_t^{(i)}\) in Eq.~\eqref{eq:direction_rectified_endpoint} serves as the rectified geometric intent of the HIST3R update, and Sec.~\ref{sec:magnitude} next determines the update gain along it.

\subsection{Modulating Update Gain via Observation Discrepancy}
\label{sec:magnitude}

After rectifying the geometric intent, HIST3R modulates the update gain, i.e., how strongly the current observation should modify the recurrent state along \(\hat{d}_t^{(i)}\). 
A suitable update gain depends on how much new evidence the observation provides relative to recent history: if the observation is already consistent with recent visual history, a large update may redundantly perturb the recurrent state; if it provides complementary geometric evidence not captured by recent history, an overly conservative update may prevent the state from adapting.

HIST3R therefore modulates the candidate update gain according to an observation discrepancy score \(r_t^{(i)}\), which measures how much the current visual context differs from its historical context:
\begin{equation}
    \beta_t^{(i)}
    =
    \sigma\!\left(
    \lambda_{\mathrm{disc}}
    \left(r_t^{(i)}-\theta_{\mathrm{disc}}\right)
    \right),
    \qquad
    \hat{m}_t^{(i)}
    =
    \beta_t^{(i)} \tilde{m}_t^{(i)},
\label{eq:rectified_magnitude}
\end{equation}
where \(\lambda_{\mathrm{disc}}\) and \(\theta_{\mathrm{disc}}\) control the sharpness and threshold of the gate. 
A small discrepancy indicates that the observation is largely explained by recent history, yielding a conservative update gain that suppresses redundant state changes, while a large discrepancy indicates new evidence and allows a stronger update along \(\hat{d}_t^{(i)}\). 
Combined with \(\hat{d}_t^{(i)}\) from Sec.~\ref{sec:direction}, \(\hat{m}_t^{(i)}\) yields the final HIST3R update in Eq.~\ref{eq:hist3r_update}. 
We next describe how the visual context is summarized and how \(r_t^{(i)}\) is computed.

\paragraph{State-token-specific visual summary.}
Computing \(r_t^{(i)}\) requires a compact summary of each frame that can be compared across time. 
A simple choice is to average-pool all image tokens into a single global token. 
However, different state tokens attend to different image regions when interacting with the frame through the decoders in Eq.~\ref{eq:state_image_interaction}, so a single global summary discards the visual evidence relevant to each state token. 
HIST3R instead summarizes each frame per state token, aggregating image tokens according to the first-layer state-to-image attention map \(A_t \in \mathbb{R}^{N \times P}\) from the decoder:
\begin{equation}
    V_t^{(i)}
    =
    \sum_{j=1}^{P}
    A_t^{(i,j)} F_t^{(j)},
\label{eq:visual_summary}
\end{equation}
where \(F_t^{(j)}\) denotes the \(j\)-th image token. 
Each \(V_t^{(i)}\) thus represents the visual evidence observed by state token \(i\). 
The resulting \(V_t \in \mathbb{R}^{N \times C}\) is stored in the history buffer \(\mathcal{H}_t\) defined in Eq.~\ref{eq:history_window}.

\paragraph{Discrepancy-guided modulation.}
The discrepancy score \(r_t^{(i)}\) measures how much additional visual evidence the current observation provides beyond relevant history. 
For each state token \(i\), HIST3R compares the current summary \(V_t^{(i)}\) with each historical summary \(V_\tau^{(i)}\) using a normalized L2 distance, and aggregates these distances with the local history weights \(w_{t,\tau}^{(i)}\) from Eq.~\ref{eq:history_weights_proposal}:
\begin{equation}
    r_{t,\tau}^{(i)}
    =
    \frac{
    \left\|V_t^{(i)} - V_\tau^{(i)}\right\|_2
    }{
    \left\|V_\tau^{(i)}\right\|_2 + \epsilon
    },
    \qquad
    r_t^{(i)}
    =
    \sum_{\tau \in \mathcal{T}_t}
    w_{t,\tau}^{(i)} r_{t,\tau}^{(i)}.
\label{eq:observation_discrepancy}
\end{equation}
Since \(w_{t,\tau}^{(i)}\) assigns larger weights to historical states that are more similar to the current candidate state, the discrepancy is measured mainly against the most relevant history rather than against the entire history uniformly. 
The resulting \(r_t^{(i)}\) determines the update gain through Eq.~\ref{eq:rectified_magnitude}.

Together with the rectified geometric intent \(\hat{d}_t^{(i)}\) from Sec.~\ref{sec:direction}, this update gain yields the final HIST3R update in Eq.~\ref{eq:hist3r_update}, which controls both components of the recurrent update using history while remaining training-free.
\section{Experiments}
\label{sec:experiments}
We evaluate HIST3R across camera pose estimation (Sec.~\ref{sec:exp_pose_estimation}), 
3D reconstruction (Sec.~\ref{sec:exp_3d_reconstruction}), and video depth estimation 
(Sec.~\ref{sec:exp_video_depth}), followed by ablations that validate the proposed 
direction--magnitude decomposition view (Sec.~\ref{sec:ablation}).

\paragraph{Baselines and protocol.}
We compare HIST3R with recent online reconstruction methods. 
CUT3R~\citep{wang2025cut3r} serves as the base implicit-memory model. 
Our main comparisons are TTT3R~\citep{chen2026ttt3r} and TTSA3R~\citep{zheng2026ttsa3r}, two training-free CUT3R extensions that modulate the update gain while leaving the geometric intent unchanged. 
We also compare with explicit-memory methods, including Point3R~\citep{wu2025point3r}, StreamVGGT~\citep{streamvggt2026}, and InfiniteVGGT~\citep{yuan2026infinitevggt}. All hyperparameters are selected once on the ScanNet training split and remain fixed across all datasets, tasks, and sequence lengths, as detailed in Appendix~\ref{app:hyperparameters}.
Since HIST3R is designed for long-horizon streaming reconstruction, the main paper focuses on long-sequence evaluation. Sparse-view and standard short-sequence results are reported in Appendix~\ref{app:additional_settings}. All methods are evaluated on a single NVIDIA H100 GPU with 80GB memory.

\subsection{Camera Pose Estimation}
\label{sec:exp_pose_estimation}

For camera pose estimation, we evaluate on ScanNet~\citep{scannet} and TUM-Dynamics~\citep{tumd}, following the protocol of prior online reconstruction methods~\citep{wang2025cut3r,chen2026ttt3r}. We use Absolute Trajectory Error (ATE) after Sim(3) trajectory alignment~\citep{tumd} as the main evaluation metric.

\begin{figure}
    \centering
    \includegraphics[width=\linewidth]{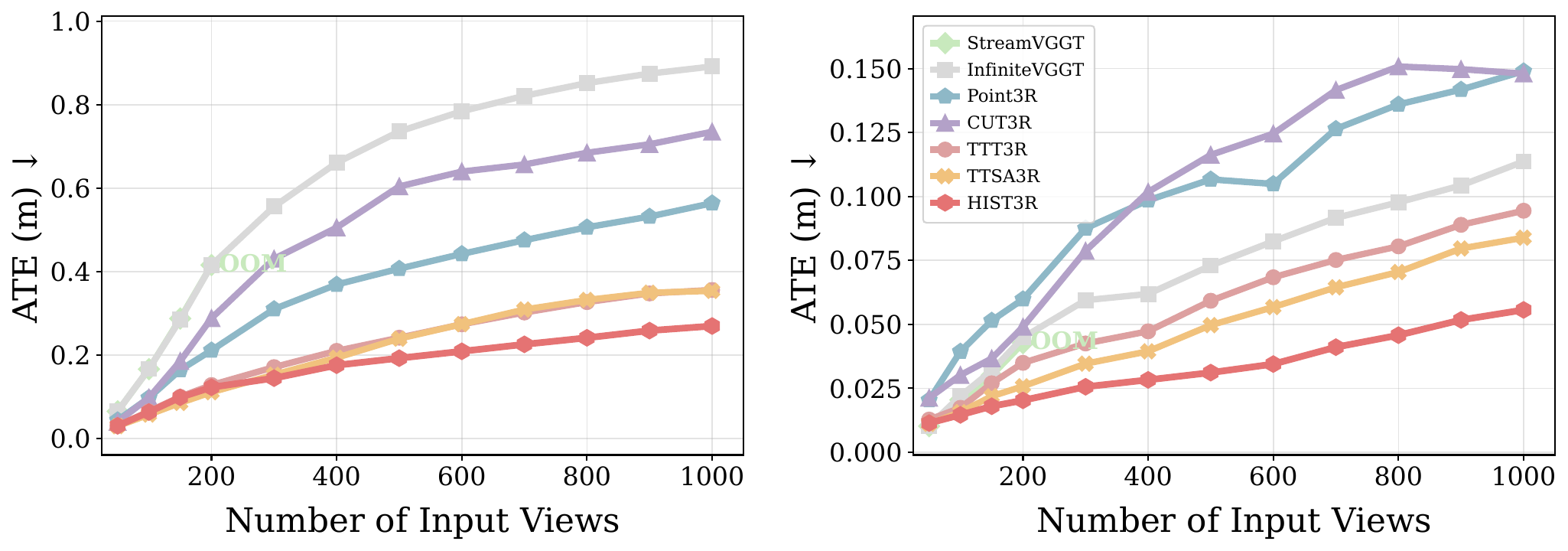}
    \caption{
        \textbf{Camera pose estimation on ScanNet~\citep{scannet} (left) and TUM-Dynamics~\citep{tumd} (right).} 
        OOM denotes out-of-memory under an 80GB GPU memory budget.
    }
    \label{fig:camera_pose}
\end{figure}


As shown in Fig.~\ref{fig:camera_pose}, HIST3R consistently achieves the lowest ATE across long input sequences on both ScanNet and TUM-Dynamics. 
The improvement becomes more pronounced as the sequence length increases: at 1000 input frames, HIST3R reduces ATE over the second-best method, TTSA3R, by \(\mathbf{23.87\%}\) on ScanNet and \(\mathbf{33.65\%}\) on TUM-Dynamics. 
This trend supports our motivation that long-horizon streaming reconstruction benefits from controlling not only the update gain but also the geometric intent.

\subsection{3D Reconstruction}
\label{sec:exp_3d_reconstruction}

\providecommand{\oom}{\textcolor{lightgray}{\textit{OOM}}}
\definecolor{methodsepgray}{gray}{0.60}

\newcommand{\methodsep}{%
    \arrayrulecolor{methodsepgray}%
    \cmidrule[0.025em](lr){1-1}\cmidrule[0.025em](lr){3-14}%
    \arrayrulecolor{black}%
}
\begin{table*}[t]
\centering
\caption{
    \textbf{3D reconstruction results on 7-Scenes~\citep{20137scenes} and NRGBD~\citep{2022NeuralRGBD}}. Best results are shown in bold, and second-best results are underlined.
}
\resizebox{\textwidth}{!}{
\begin{tabular}{llcccccccccccc}
    \toprule[0.17em]
    \multirow{3}{*}{\textbf{Method}} &
    \multirow{3}{*}{\textbf{Input}} &
    \multicolumn{6}{c}{\textbf{7-Scenes}} &
    \multicolumn{6}{c}{\textbf{NRGBD}} \\
    \cmidrule(lr){3-8} \cmidrule(lr){9-14}
    & &
    \multicolumn{2}{c}{\textbf{Acc. $\downarrow$}} &
    \multicolumn{2}{c}{\textbf{Comp. $\downarrow$}} &
    \multicolumn{2}{c}{\textbf{NC $\uparrow$}} &
    \multicolumn{2}{c}{\textbf{Acc. $\downarrow$}} &
    \multicolumn{2}{c}{\textbf{Comp. $\downarrow$}} &
    \multicolumn{2}{c}{\textbf{NC $\uparrow$}} \\
    \cmidrule(lr){3-4} \cmidrule(lr){5-6} \cmidrule(lr){7-8}
    \cmidrule(lr){9-10} \cmidrule(lr){11-12} \cmidrule(lr){13-14}
    & &
    \textbf{Mean} & \textbf{Med.} &
    \textbf{Mean} & \textbf{Med.} &
    \textbf{Mean} & \textbf{Med.} &
    \textbf{Mean} & \textbf{Med.} &
    \textbf{Mean} & \textbf{Med.} &
    \textbf{Mean} & \textbf{Med.} \\
    \midrule[0.08em]

    StreamVGGT   & \multirow{7}{*}{\textit{300}} & \oom & \oom & \oom & \oom & \oom & \oom & \oom & \oom & \oom & \oom & \oom & \oom \\
    InfiniteVGGT &  & 0.042 & 0.017 & 0.026 & 0.007 & \textbf{0.570} & \textbf{0.607} & \textbf{0.054} & \underline{0.031} & 0.025 & \underline{0.005} & \textbf{0.645} & \textbf{0.748} \\
    Point3R      &  & 0.046 & 0.025 & 0.033 & 0.014 & 0.564 & 0.597 & \underline{0.073} & 0.040 & \underline{0.013} & \textbf{0.004} & 0.614 & 0.687 \\
    \methodsep
    CUT3R        &  & 0.126 & 0.085 & 0.070 & 0.030 & 0.543 & 0.564 & 0.215 & 0.118 & 0.064 & 0.013 & 0.570 & 0.607 \\
    TTT3R        &  & 0.041 & 0.025 & 0.024 & \underline{0.005} & 0.565 & 0.599 & 0.101 & 0.040 & 0.022 & \textbf{0.004} & 0.607 & 0.675 \\
    TTSA3R       &  & \underline{0.029} & \underline{0.015} & \underline{0.022} & \textbf{0.004} & \underline{0.567} & \underline{0.603} & 0.090 & 0.037 & 0.020 & \underline{0.005} & 0.613 & 0.684 \\
    \textbf{HIST3R} &  & \textbf{0.020} & \textbf{0.009} & \textbf{0.021} & \textbf{0.004} & 0.565 & 0.599 & 0.075 & \textbf{0.025} & \textbf{0.012} & \textbf{0.004} & \underline{0.620} & \underline{0.695} \\

    \midrule[0.08em]

    StreamVGGT   & \multirow{7}{*}{\textit{400}} & \oom & \oom & \oom & \oom & \oom & \oom & \oom & \oom & \oom & \oom & \oom & \oom \\
    InfiniteVGGT &  & 0.040 & \underline{0.015} & 0.027 & 0.006 & \textbf{0.565} & \textbf{0.599} & \textbf{0.066} & \underline{0.035} & \underline{0.035} & 0.007 & \textbf{0.647} & \textbf{0.752} \\
    Point3R      &  & 0.053 & 0.026 & \underline{0.025} & 0.009 & 0.560 & 0.591 & 0.090 & 0.043 & \textbf{0.024} & \textbf{0.004} & 0.610 & 0.681 \\
    \methodsep
    CUT3R        &  & 0.152 & 0.102 & 0.094 & 0.041 & 0.536 & 0.552 & 0.305 & 0.220 & 0.108 & 0.038 & 0.554 & 0.580 \\
    TTT3R        &  & 0.052 & 0.031 & 0.027 & \underline{0.005} & 0.557 & 0.586 & 0.142 & 0.066 & 0.053 & 0.008 & 0.593 & 0.647 \\
    TTSA3R       &  & \underline{0.035} & 0.019 & \textbf{0.023} & \textbf{0.004} & 0.560 & 0.591 & 0.103 & 0.046 & \underline{0.035} & \underline{0.006} & 0.608 & 0.674 \\
    \textbf{HIST3R} &  & \textbf{0.021} & \textbf{0.010} & \textbf{0.023} & \textbf{0.004} & \underline{0.561} & \underline{0.592} & \underline{0.084} & \textbf{0.034} & \textbf{0.024} & \textbf{0.004} & \underline{0.616} & \underline{0.688} \\

    \midrule[0.08em]

    StreamVGGT   & \multirow{7}{*}{\textit{500}} & \oom & \oom & \oom & \oom & \oom & \oom & \oom & \oom & \oom & \oom & \oom & \oom \\
    InfiniteVGGT &  & \underline{0.038} & \underline{0.014} & \underline{0.023} & \underline{0.005} & \textbf{0.562} & \textbf{0.593} & \textbf{0.070} & \underline{0.046} & 0.037 & 0.008 & \textbf{0.641} & \textbf{0.741} \\
    Point3R      &  & 0.054 & 0.026 & 0.025 & 0.010 & 0.555 & 0.583 & 0.114 & 0.049 & \textbf{0.026} & \textbf{0.004} & \underline{0.611} & \underline{0.683} \\
    \methodsep
    CUT3R        &  & 0.179 & 0.126 & 0.093 & 0.028 & 0.530 & 0.544 & 0.329 & 0.244 & 0.151 & 0.050 & 0.549 & 0.571 \\
    TTT3R        &  & 0.066 & 0.039 & 0.029 & 0.006 & 0.550 & 0.575 & 0.166 & 0.091 & 0.065 & 0.008 & 0.584 & 0.633 \\
    TTSA3R       &  & 0.045 & 0.023 & 0.025 & \textbf{0.004} & 0.555 & 0.583 & 0.120 & 0.053 & 0.050 & \underline{0.006} & 0.604 & 0.667 \\
    \textbf{HIST3R} &  & \textbf{0.023} & \textbf{0.010} & \textbf{0.022} & \textbf{0.004} & \underline{0.557} & \underline{0.586} & \underline{0.092} & \textbf{0.035} & \underline{0.033} & \textbf{0.004} & 0.610 & 0.678 \\

    \bottomrule[0.17em]
\end{tabular}
}
\label{tab:mv-recon-all}
\end{table*}
Following prior work~\citep{wang2025cut3r,wu2025point3r,yuan2026infinitevggt}, we evaluate 3D reconstruction on 7-Scenes~\citep{20137scenes} and NRGBD~\citep{2022NeuralRGBD}. 
We assess reconstruction quality by comparing predicted pointmaps with ground-truth point clouds using accuracy, completeness, and normal consistency (NC). 
Accuracy measures the nearest-neighbor distance from the predicted reconstruction to the ground truth, while completeness measures the distance in the reverse direction; for both metrics, we report mean and median values. 
To evaluate long-sequence reconstruction, we test each method on input sequences of 300, 400, and 500 frames.

As shown in Table~\ref{tab:mv-recon-all}, HIST3R achieves the strongest reconstruction quality on 7-Scenes, obtaining the best accuracy and completeness across long input sequences while remaining competitive in normal consistency. 
On NRGBD, HIST3R achieves the best or second-best results on most metrics. 
Compared with explicit-memory methods such as InfiniteVGGT and Point3R, no single method dominates every metric, and HIST3R remains competitive in accuracy and completeness. 
Its main advantage lies in efficiency: by maintaining a compact recurrent state, HIST3R requires substantially less memory and runs considerably faster than explicit-memory methods, as analyzed in Appendix~\ref{app:efficiency}.

The advantage is especially clear within the CUT3R-based implicit-memory family. 
HIST3R consistently outperforms CUT3R, TTT3R, and TTSA3R across nearly all datasets, sequence lengths, and reconstruction metrics. 
At 500 frames, HIST3R improves median accuracy over TTSA3R, the second-best CUT3R variant, by \(\mathbf{56.52\%}\) on 7-Scenes and \(\mathbf{33.96\%}\) on NRGBD. 
This margin indicates that controlling both geometric intent and update gain substantially strengthens CUT3R's long-horizon reconstruction capability.

\subsection{Video Depth Estimation}
\label{sec:exp_video_depth}
\begin{figure}[t]
    \centering
    \includegraphics[width=\linewidth]{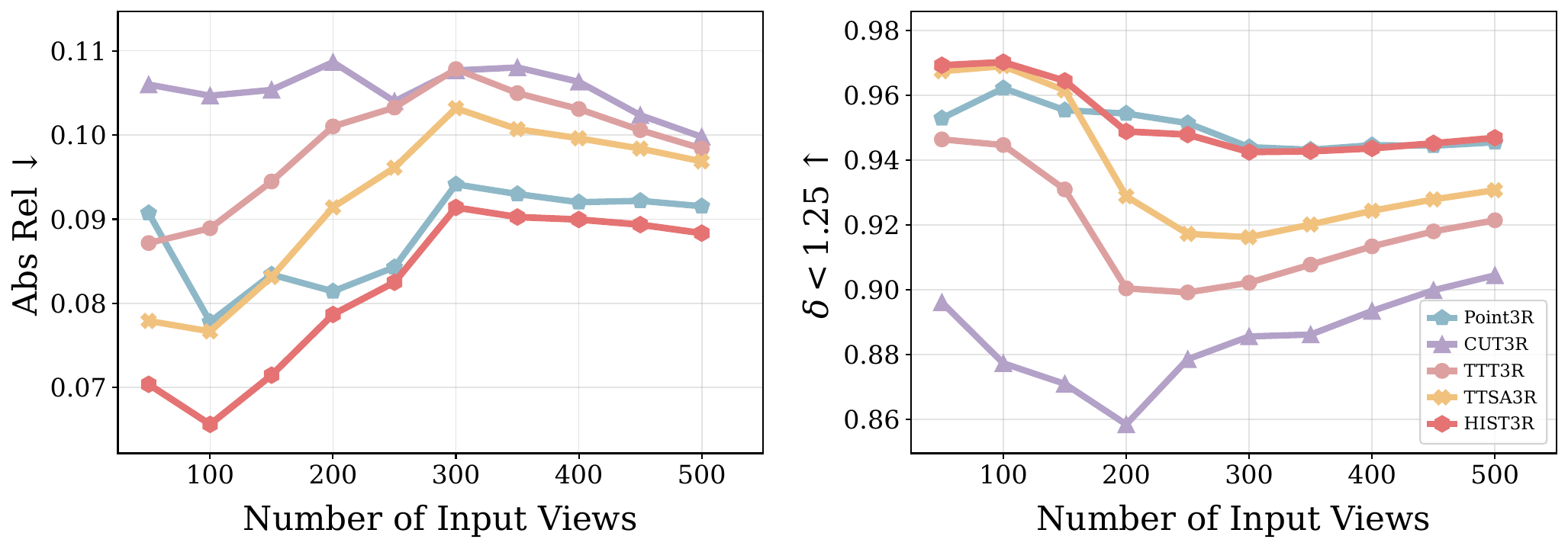}
    \caption{
        \textbf{Metric depth evaluation on Bonn~\citep{bonn}.}
        We exclude VGGT-based methods that do not support metric depth prediction.
    }
    \label{fig:bonn_depth}
\end{figure}

\begin{figure}[t]
    \centering
    \includegraphics[width=\linewidth]{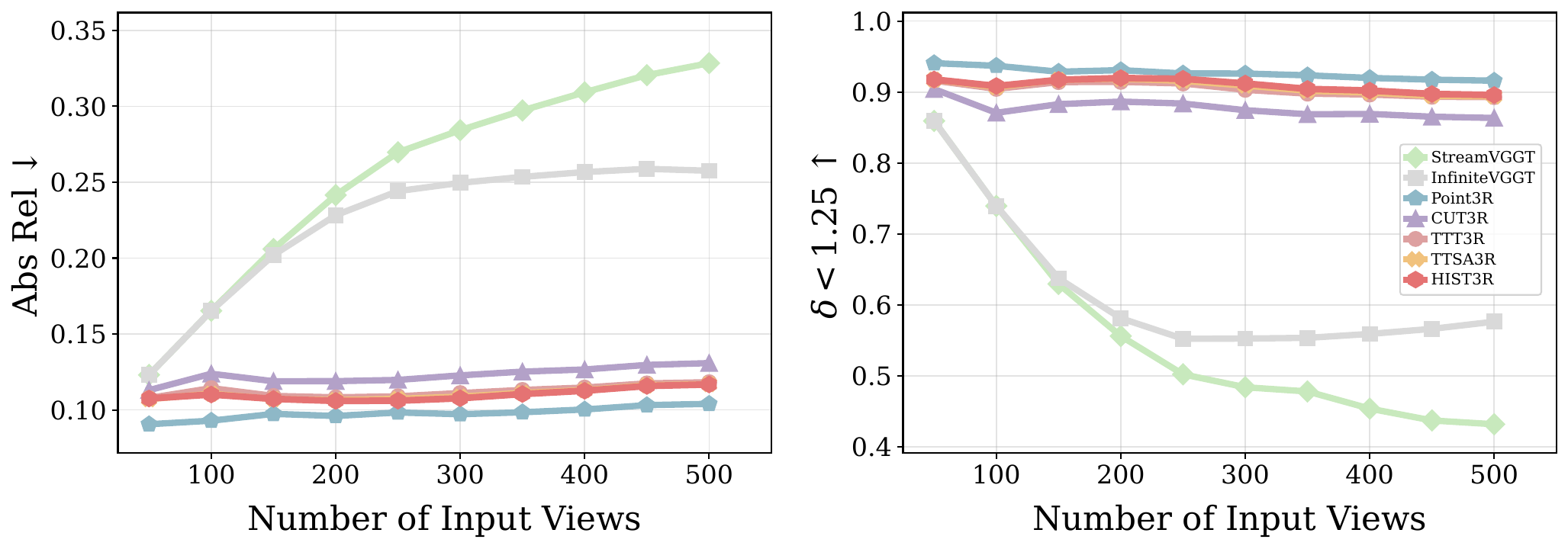}
    \caption{
        \textbf{Scale-aligned depth evaluation on KITTI~\citep{kitti}.}
    }
    \label{fig:kitti_depth}
\end{figure}
Following prior online reconstruction work~\citep{wang2025cut3r,chen2026ttt3r}, we evaluate video depth estimation on KITTI~\citep{kitti} and Bonn~\citep{bonn}, which cover outdoor driving and dynamic indoor scenes. We report absolute relative error (AbsRel) and \(\delta < 1.25\), the percentage of predicted depths within a factor of \(1.25\) of the ground truth, under two protocols: scale-aligned evaluation, which aligns each predicted sequence to the ground truth with a per-sequence scale to measure relative depth quality, and metric-depth evaluation, which compares predictions in metric units without alignment and is reported only for methods that output metric-scale pointmaps. 

As shown in Fig.~\ref{fig:bonn_depth}, HIST3R achieves the strongest metric-depth performance on Bonn across both AbsRel and \(\delta < 1.25\), especially under longer input sequences. 
On KITTI, HIST3R ranks second behind Point3R in scale-aligned depth evaluation and remains competitive, as shown in Fig.~\ref{fig:kitti_depth}.

\subsection{Ablation Studies}
\label{sec:ablation}

We ablate the two update controls introduced by the direction--magnitude decomposition, as well as the design choices within each.

\paragraph{Component ablation.}
\begin{table*}
\centering
\scriptsize
\renewcommand{\arraystretch}{0.98}
\setlength{\tabcolsep}{3.2pt}
\caption{
    \textbf{Ablation of direction--magnitude decomposition.}
    We evaluate direction rectification, magnitude modulation, and their combination on camera pose estimation, scale-aligned video depth estimation, and 3D reconstruction, all under the 300-frame setting.
    AR denotes AbsRel, \(\delta\) denotes \(\delta < 1.25\), and reconstruction Acc./Comp. are reported as mean accuracy/completeness.
}
\resizebox{\textwidth}{!}{
\begin{tabular}{ll c cc cc cc cc}
    \toprule[0.14em]
    \multirow{2}{*}{\textbf{Method}} &
    \multirow{2}{*}{\textbf{Update Rule}} &
    \textbf{TUM-D} &
    \multicolumn{2}{c}{\textbf{Bonn}} &
    \multicolumn{2}{c}{\textbf{KITTI}} &
    \multicolumn{2}{c}{\textbf{7-Scenes}} &
    \multicolumn{2}{c}{\textbf{NRGBD}} \\
    \cmidrule(lr){3-3}
    \cmidrule(lr){4-5}
    \cmidrule(lr){6-7}
    \cmidrule(lr){8-9}
    \cmidrule(lr){10-11}
    &
    &
    \textbf{ATE $\downarrow$} &
    \textbf{AR $\downarrow$} &
    \textbf{$\delta \uparrow$} &
    \textbf{AR $\downarrow$} &
    \textbf{$\delta \uparrow$} &
    \textbf{Acc. $\downarrow$} &
    \textbf{Comp. $\downarrow$} &
    \textbf{Acc. $\downarrow$} &
    \textbf{Comp. $\downarrow$} \\
    \midrule[0.06em]

    CUT3R baseline
    & \(S_t=S_{t-1}+\widetilde{\mathbf{m}}_t\widetilde{\mathbf{d}}_t\)
    & 0.079 & 0.089 & 0.938 & 0.123 & 0.874 & 0.126 & 0.070 & 0.215 & 0.064 \\
    + Direction rect.
    & \(S_t=S_{t-1}+\widetilde{\mathbf{m}}_t\widehat{\mathbf{d}}_t\)
    & 0.032 & \textbf{0.077} & \underline{0.952} & \underline{0.109} & 0.907 & 0.027 & \underline{0.023} & 0.086 & \underline{0.017} \\

    + Magnitude mod.
    & \(S_t=S_{t-1}+\widehat{\mathbf{m}}_t\widetilde{\mathbf{d}}_t\)
    & \underline{0.027} & \underline{0.080} & \underline{0.952} & \underline{0.109} & \underline{0.909} & \underline{0.021} & \textbf{0.021} & \underline{0.077} & \textbf{0.012} \\

    \textbf{HIST3R}
    & \(S_t=S_{t-1}+\widehat{\mathbf{m}}_t\widehat{\mathbf{d}}_t\)
    & \textbf{0.026} & \textbf{0.077} & \textbf{0.954} & \textbf{0.108} & \textbf{0.912} & \textbf{0.020} & \textbf{0.021} & \textbf{0.075} & \textbf{0.012} \\

    \bottomrule[0.14em]
\end{tabular}
}

\label{tab:component_selected}
\end{table*}
We isolate the two update controls by applying only direction rectification or only magnitude modulation. 
As shown in Table~\ref{tab:component_selected}, both improve over CUT3R across camera pose estimation, video depth estimation, and 3D reconstruction, and their combination achieves the best overall performance. 
This supports the decomposition view in Eq.~\ref{eq:hist3r_update}: reliable long-horizon updating benefits from jointly rectifying the geometric intent and modulating the update gain.

\paragraph{Design ablations.}
\begin{table*}
\centering
\captionsetup{skip=6pt}
\caption{
    \textbf{Design ablations} of direction rectification (left) and magnitude modulation (right), evaluated under the 200-frame setting with scale-aligned depth.
    The check marks indicate whether cosine-weighted aggregation, the alignment gate \(\alpha_t\), and direction-only rectification are applied.
}
\label{tab:design-ablation}

\begin{minipage}[t]{0.54\textwidth}
\centering
\resizebox{\linewidth}{!}{
\begin{tabular}{lcccccc}
    \toprule[0.17em]
    \multicolumn{7}{c}{
        \textbf{Direction Rectification:}
        $S_t = S_{t-1} + \widetilde{\mathbf{m}}_t \, \widehat{\mathbf{d}}_t$
    } \\
    \midrule[0.08em]
    \textbf{Design} &
    \textbf{Cos. Agg.} &
    \textbf{Align. Gate} &
    \textbf{Dir. Only} &
    \textbf{TUM ATE $\downarrow$} &
    \textbf{KITTI AR $\downarrow$} &
    \textbf{KITTI $\delta \uparrow$} \\
    \midrule[0.08em]
    \multirow{4}{*}{
    \shortstack[c]{
        Protocol \\
        {\small\textit{($k=8$)}}
    }
    }
    & \yesmark & \yesmark & \yesmark & \textbf{0.024} & \textbf{0.107} & \textbf{0.917} \\
    & \nomark  & \yesmark & \yesmark & \underline{0.026} & \underline{0.110} & \underline{0.895} \\
    & \yesmark & \nomark  & \yesmark & 0.037 & 0.119 & 0.871 \\
    & \yesmark & \yesmark & \nomark  & 0.028 & 0.123 & 0.863 \\
    \midrule[0.08em]
    \multirow{6}{*}{Window size \(k\)}
    & \multicolumn{3}{c}{\(k=2\)}  & 0.044 & 0.114 & 0.896 \\
    & \multicolumn{3}{c}{\(k=4\)}  & 0.035 & 0.110 & 0.908 \\
    & \multicolumn{3}{c}{\(k=8\) {\small\textit{(default)}}} & 0.024 & \textbf{0.107} & \textbf{0.917} \\
    & \multicolumn{3}{c}{\(k=16\)} & \textbf{0.021} & \underline{0.108} & \underline{0.916} \\
    & \multicolumn{3}{c}{\(k=32\)} & \textbf{0.021} & \underline{0.108} & \underline{0.916} \\
    & \multicolumn{3}{c}{\(k=64\)} & \underline{0.023} & 0.110 & 0.913 \\
    \bottomrule[0.17em]
\end{tabular}
}
\end{minipage}
\hfill
\begin{minipage}[t]{0.44\textwidth}
\centering
\resizebox{\linewidth}{!}{
\begin{tabular}{llccc}
    \toprule[0.17em]
    \multicolumn{5}{c}{
        \textbf{Magnitude Modulation:}
        $S_t = S_{t-1} + \widehat{\mathbf{m}}_t \, \widetilde{\mathbf{d}}_t$
    } \\
    \midrule[0.08em]
    \textbf{Design} &
    \textbf{Attn. Layers} &
    \textbf{TUM ATE $\downarrow$} &
    \textbf{KITTI AR $\downarrow$} &
    \textbf{KITTI $\delta \uparrow$} \\
    \midrule[0.08em]
    Avg. pooling & -- & 0.043 & 0.112 & 0.903 \\
    \midrule[0.08em]
    \multirow{4}{*}{Attn. pooling}
    & \(n=1\) {\small\textit{(default)}}  & \textbf{0.021} & \textbf{0.106} & 0.918 \\
    & \(n=3\)  & \underline{0.023} & \textbf{0.106} & \underline{0.919} \\
    & \(n=6\)  & \underline{0.023} & \textbf{0.106} & \textbf{0.920} \\
    & \(n=12\) & \underline{0.023} & \textbf{0.106} & \underline{0.919} \\
    \bottomrule[0.17em]
\end{tabular}
}
\end{minipage}
\end{table*}
Table~\ref{tab:design-ablation} ablates the design choices within each component, applying direction rectification and magnitude modulation separately.
Direction rectification consists of three stages: aggregating historical states with cosine weights to form the history-proposed endpoint in Eq.~\ref{eq:history_weights_proposal}, fusing it with the candidate endpoint through the alignment gate in Eq.~\ref{eq:direction_gate}, and using the rectified endpoint only to set the update direction in Eq.~\ref{eq:direction_rectified_endpoint}. 
Removing any of these stages degrades performance. 
Replacing cosine weights with uniform averaging weakens the history proposal, and removing the alignment gate, which imposes history unconditionally, causes the largest drop. 
Directly adopting the rectified endpoint as the new state, which alters both direction and magnitude, also degrades performance, supporting the separate control of the two components. 
For the history window, performance improves as \(k\) increases and remains stable for \(8 \le k \le 32\), while an overly long window introduces less relevant states; we therefore use \(k=8\) as the smallest stable window. 

For magnitude modulation, state-token-specific attention pooling clearly outperforms global average pooling, confirming that different state tokens rely on different image regions. 
Using only the first-layer attention map performs on par with aggregating multiple layers while allowing the remaining layers to keep FlashAttention~\citep{flashattention}, which keeps HIST3R nearly as fast as CUT3R, as shown in Appendix~\ref{app:efficiency}.

\section{Conclusion}
We presented HIST3R, a training-free framework that improves long-sequence streaming 3D reconstruction on top of CUT3R. 
The key idea is to view recurrent state updates as controllable state-space evolution and to decompose each update into direction and magnitude, which correspond to its geometric intent and update gain. 
This decomposition unifies CUT3R and its training-free variants, and reveals that existing variants control only the update gain, leaving the geometric intent as a missing control axis. 
HIST3R controls both components with history: it rectifies the geometric intent using the local structure of recent recurrent states, and modulates the update gain using observation--history visual discrepancy. 
Together, these two controls allow the compact recurrent state to incorporate useful new evidence while avoiding redundant or unstable updates over long sequences. 
Experiments across camera pose estimation, 3D reconstruction, and video depth estimation show that HIST3R consistently improves long-horizon performance while preserving the memory and speed of CUT3R. 
Overall, our results suggest that controlling both where and how strongly the recurrent state is updated provides an effective principle for robust long-horizon 3D perception.


\begin{ack}
This work is supported in part by the National Science and Technology Council (NSTC), Taiwan, under grants \mbox{NSTC 114-2221-E-002-056-MY2} and \mbox{NSTC 114-2634-F-002-006}. We also thank the National Center for High-performance Computing (NCHC) for providing computational and storage resources.
\end{ack}

\bibliographystyle{unsrtnat}
\newpage
\bibliography{references}

\newpage
\appendix

\section*{Appendix}

\section{More Experimental Analysis}
\label{app:additional_experiments}
We provide additional analyses and results in this section. 
We first analyze the local structure of CUT3R state evolution in Appendix~\ref{app:state_structure}, which motivates our history-guided direction rectification. 
We then detail the hyperparameter selection protocol and sensitivity of HIST3R in Appendix~\ref{app:hyperparameters}, and show in Appendix~\ref{app:plugin} that direction rectification can serve as a plug-in component for existing magnitude-based methods. 
Finally, we compare the streaming efficiency of HIST3R with existing methods in Appendix~\ref{app:efficiency}.

\subsection{Local Structure in CUT3R State Evolution}
\label{app:state_structure}
\begin{figure}[!htbp]
    \centering

    \begin{subfigure}{0.48\linewidth}
        \centering
        \includegraphics[width=\linewidth]{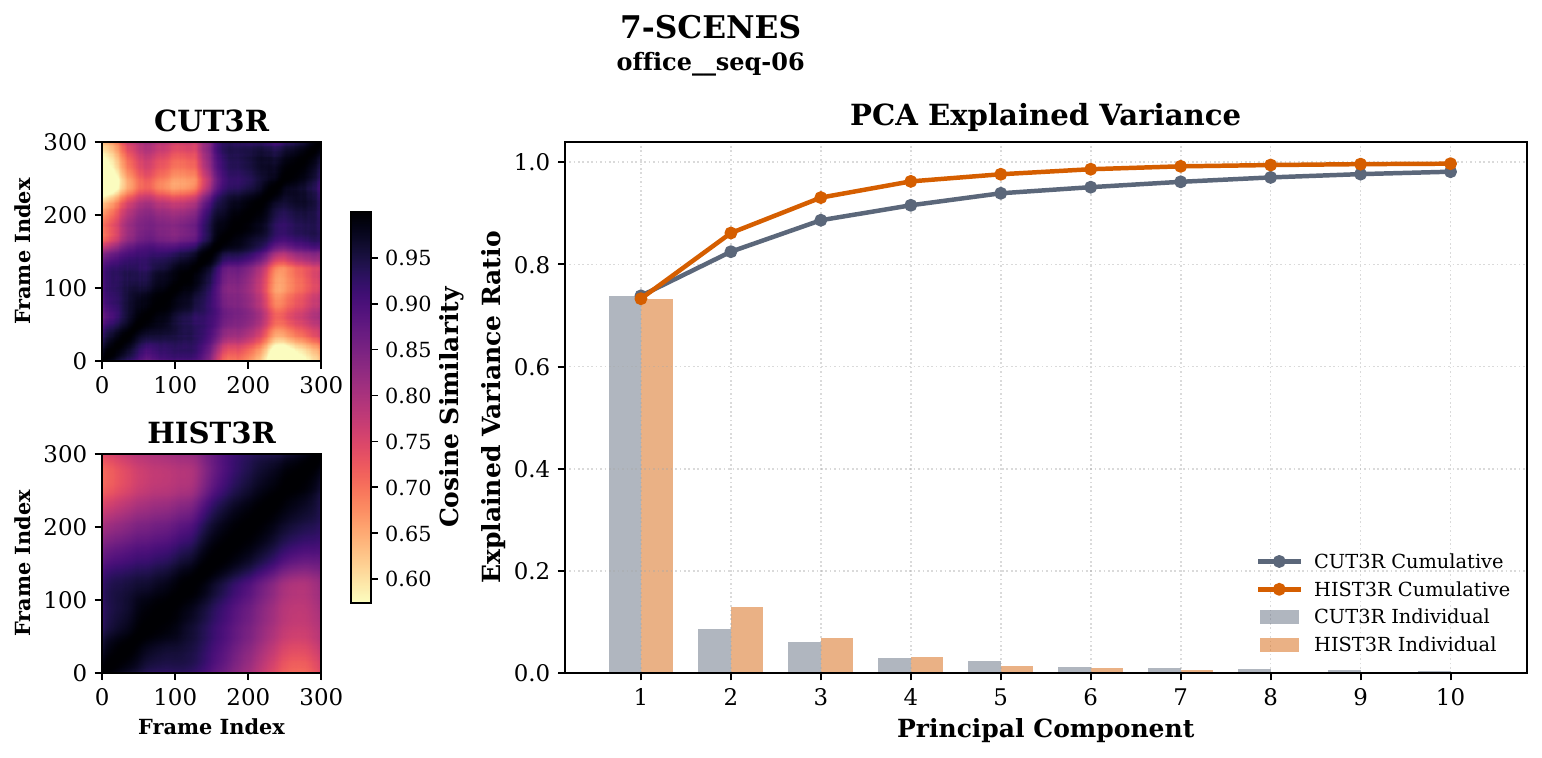}
        \caption{7-Scenes}
    \end{subfigure}
    \hfill
    \begin{subfigure}{0.48\linewidth}
        \centering
        \includegraphics[width=\linewidth]{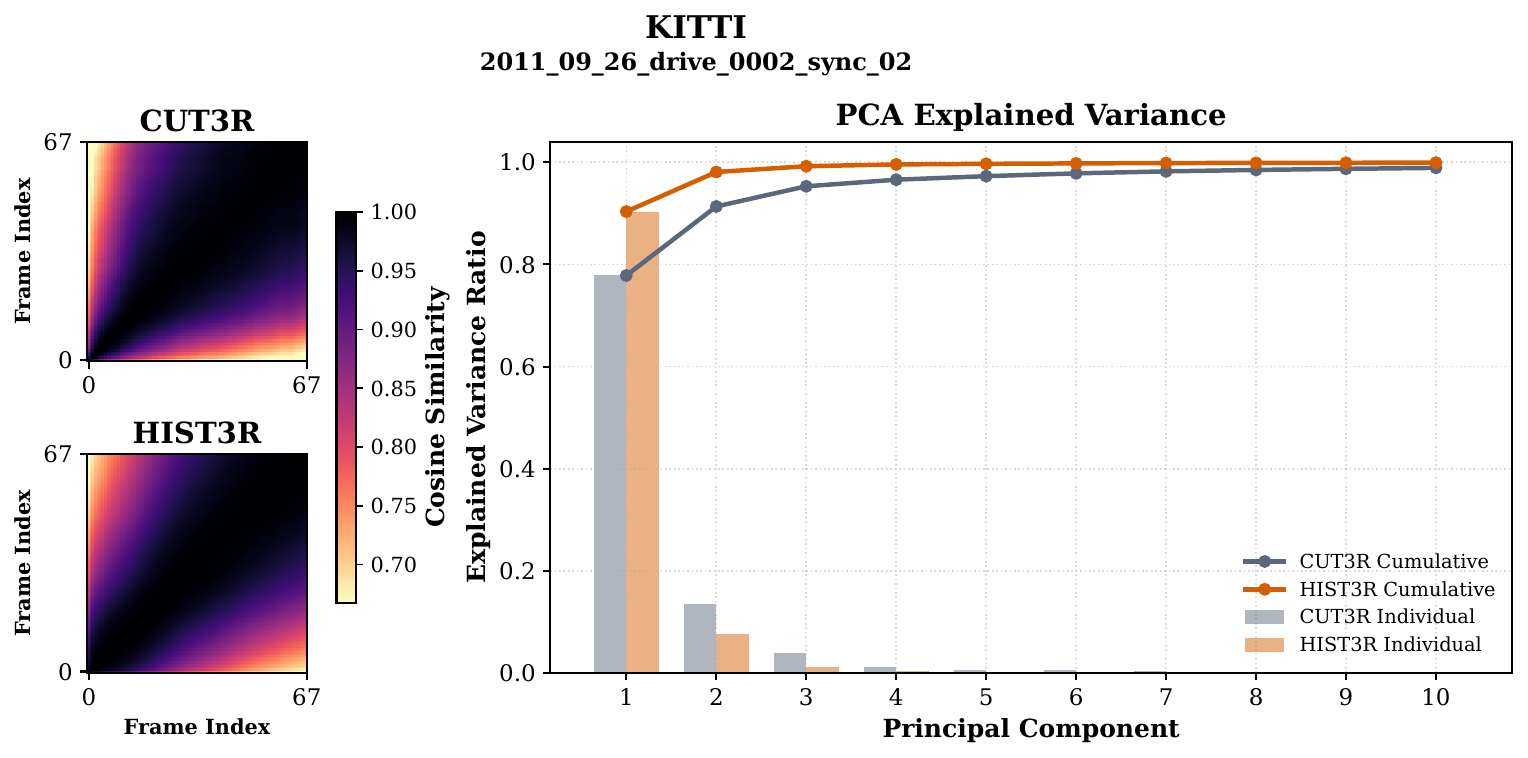}
        \caption{KITTI}
    \end{subfigure}

    \vspace{4pt}

    \begin{subfigure}{0.48\linewidth}
        \centering
        \includegraphics[width=\linewidth]{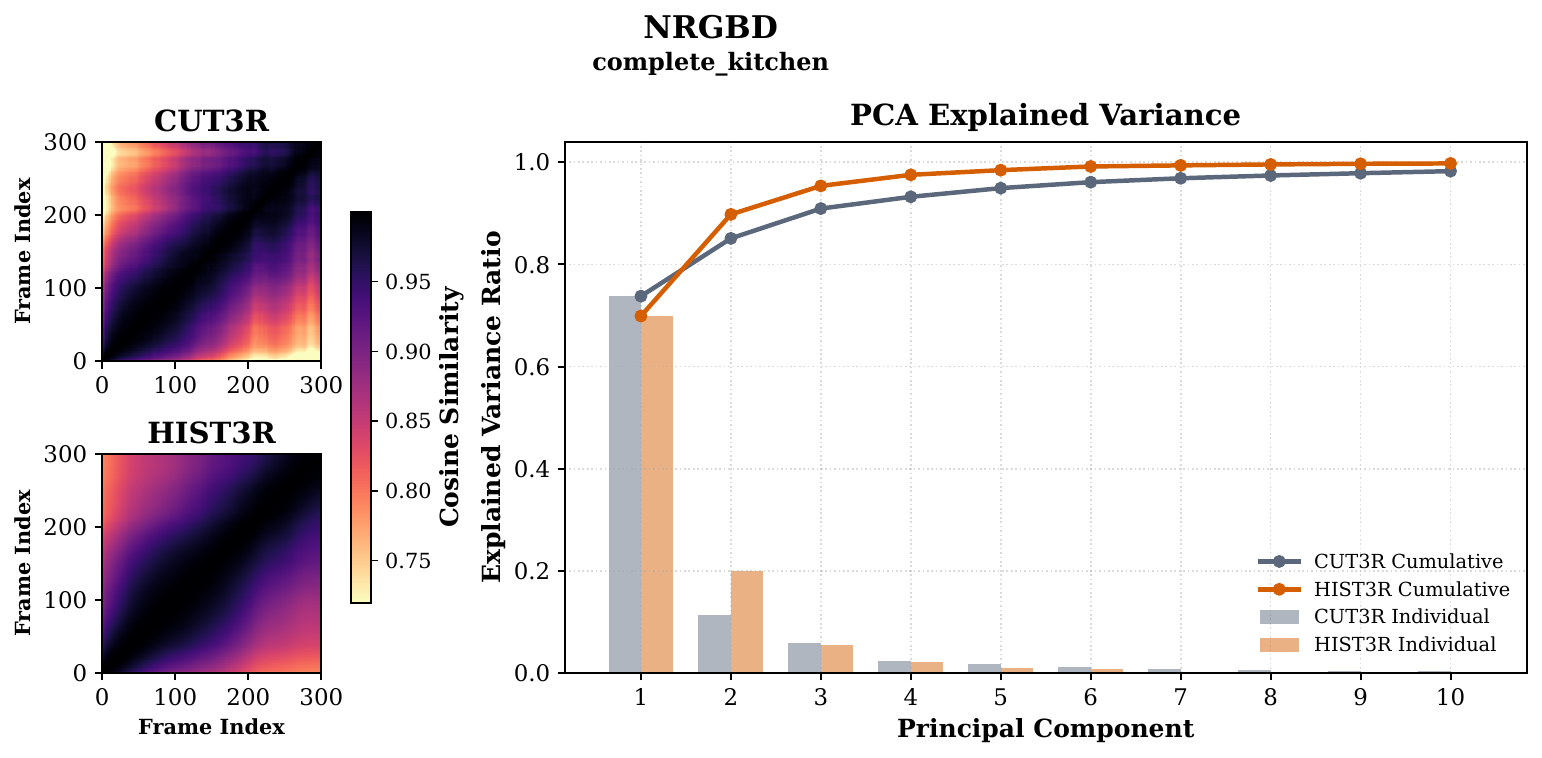}
        \caption{NRGBD}
    \end{subfigure}
    \hfill
    \begin{subfigure}{0.48\linewidth}
        \centering
        \includegraphics[width=\linewidth]{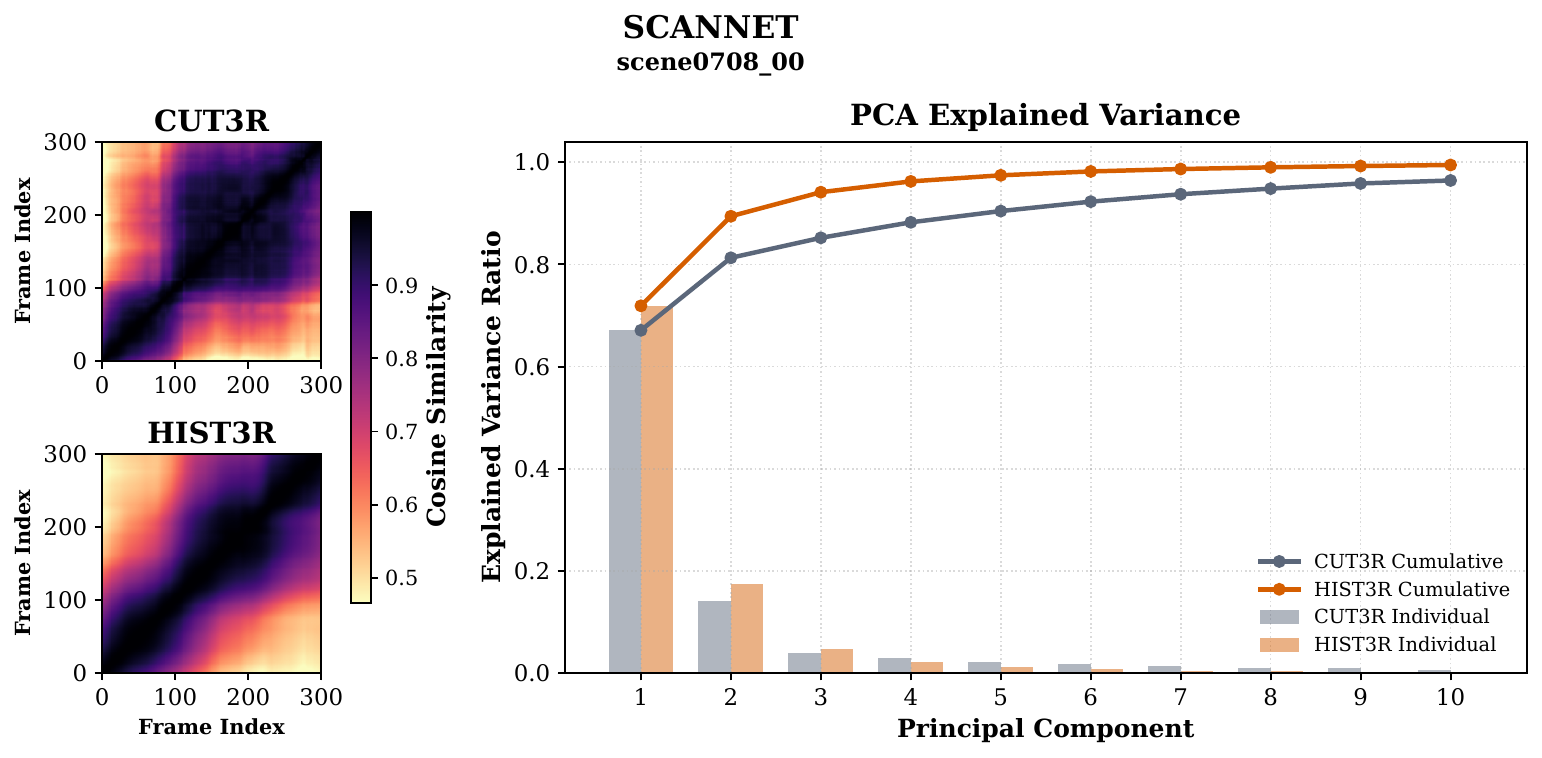}
        \caption{ScanNet}
    \end{subfigure}

    \caption{
        \textbf{Empirical local structure in CUT3R and HIST3R state evolution.}
        We visualize recurrent state evolution on one representative sequence from each of 7-Scenes~\citep{20137scenes}, KITTI~\citep{kitti}, NRGBD~\citep{2022NeuralRGBD}, and ScanNet~\citep{scannet}.
        In each subplot, the left panel shows the cosine affinity between recurrent states over time, while the right panel shows PCA analysis of state variations.
        CUT3R recurrent states naturally exhibit strong local affinity and low-dimensional variation, and HIST3R further accentuates this structure through history-guided rectification.
    }
    \label{fig:state_structure_analysis}
    \vspace{-6pt}
\end{figure}
To further support the local-structure assumption used in HIST3R, we empirically analyze the recurrent state evolution of CUT3R during streaming inference. Given a sequence of recurrent states, we examine two complementary properties: the temporal affinity between states and the intrinsic dimensionality of their variations. Specifically, we compute the cosine affinity between recurrent states over time and apply PCA to the corresponding state differences. As shown in Fig.~\ref{fig:state_structure_analysis}, representative sequences from 7-Scenes~\citep{20137scenes}, KITTI~\citep{kitti}, NRGBD~\citep{2022NeuralRGBD}, and ScanNet~\citep{scannet} consistently exhibit strong local structure. Adjacent CUT3R recurrent states have high cosine affinity, indicating that state evolution changes smoothly over short temporal windows rather than following arbitrary frame-wise transitions. Moreover, the PCA results show that state variations are largely concentrated in a small number of principal directions, suggesting that recent states trace a compact local state manifold. These observations provide empirical evidence for using recent history as a local reference for rectifying recurrent updates. In particular, when the CUT3R candidate endpoint is compatible with the local evolution pattern, it can be approximated by locally similar historical states, consistent with the self-expressiveness principle in low-dimensional subspace modeling~\citep{elhamifar2013sparse}. HIST3R builds on this property by constructing a history-proposed endpoint from recent states and applying alignment-aware rectification, which further emphasizes the local structure already present in CUT3R state evolution.

\subsection{Hyperparameter Selection and Sensitivity}
\label{app:hyperparameters}

\paragraph{Selection protocol.}
All hyperparameters of HIST3R are selected once on 30 randomly sampled sequences from the ScanNet training split and kept fixed across all datasets, tasks, and sequence lengths. 
No test data or test performance is used for parameter selection. 
The final configuration uses a history window size of \(k=8\) and \((\lambda_{\mathrm{align}}, \theta_{\mathrm{align}}, \lambda_{\mathrm{disc}}, \theta_{\mathrm{disc}})=(2, 0.94, 10, 0.24)\) throughout all experiments. 
The choice of \(k\) is analyzed in the design ablations in Sec.~\ref{sec:ablation}, where performance remains stable for \(8 \le k \le 32\).

\paragraph{Sensitivity analysis.}
\begin{table}[t]
\centering
\caption{
    \textbf{Hyperparameter sensitivity.}
    Mean and standard deviation over all tested gate configurations of each component. 
    For the last row, direction rectification uses the default setting while magnitude modulation varies. 
    TUM-Dynamics ATE is evaluated at 1000 frames and KITTI scale-aligned AbsRel at 500 frames.
}
\label{tab:sensitivity}
\small
\begin{tabular}{lcc}
\toprule
Method & TUM ATE \(\downarrow\) & KITTI AbsRel \(\downarrow\) \\
\midrule
CUT3R & 0.148 & 0.131 \\
TTT3R & 0.094 & 0.118 \\
TTSA3R & 0.084 & 0.117 \\
\midrule
Direction rect. & 0.067 \(\pm\) 0.008 & 0.118 \(\pm\) 0.000 \\
Magnitude mod. & 0.093 \(\pm\) 0.025 & 0.120 \(\pm\) 0.001 \\
Direction rect. (default) + Magnitude mod. & 0.066 \(\pm\) 0.009 & 0.117 \(\pm\) 0.001 \\
\bottomrule
\end{tabular}
\end{table}
To verify that HIST3R does not rely on narrowly tuned gate parameters, we evaluate a broad range of configurations for each component. 
For direction rectification, we vary the gate sharpness \(\lambda_{\mathrm{align}} \in \{1, 2\}\) and threshold \(\theta_{\mathrm{align}} \in \{0.5, 0.6, 0.7, 0.8, 0.9\}\), resulting in 11 configurations including the default. 
For magnitude modulation, we vary \(\lambda_{\mathrm{disc}} \in \{1, 2, 10\}\) and \(\theta_{\mathrm{disc}} \in \{0.1, 0.2, 0.3, 0.4\}\), resulting in 13 configurations including the default. 
As shown in Table~\ref{tab:sensitivity}, direction rectification is stable across all tested configurations and consistently outperforms CUT3R, TTT3R, and TTSA3R on 1000-frame TUM-Dynamics. 
Magnitude modulation shows larger variation on this long-horizon setting, but adding the default direction rectification reduces both the mean error and its variation, from \(0.093 \pm 0.025\) to \(0.066 \pm 0.009\). 
These results indicate that the gains of HIST3R do not depend on specific gate parameters, and that direction rectification also stabilizes long-horizon updates.

\subsection{Direction Rectification as a Plug-in Component}
\label{app:plugin}
\begin{table}[t]
\centering
\caption{
    \textbf{Direction rectification as a plug-in component.}
    We replace only the update direction of TTT3R and TTSA3R with the rectified direction of HIST3R while keeping their original magnitude controllers. 
    ScanNet ATE is evaluated at 1000 frames and Bonn metric-depth AbsRel at 300 frames.
}
\label{tab:plugin}
\small
\begin{tabular}{lcc}
\toprule
Method & ScanNet ATE \(\downarrow\) & Bonn AbsRel \(\downarrow\) \\
\midrule
Direction rect. only & 0.354 & 0.104 \\
\midrule
TTT3R & 0.356 & 0.108 \\
TTT3R + Direction rect. & \textbf{0.322} & \textbf{0.103} \\
\midrule
TTSA3R & 0.354 & 0.103 \\
TTSA3R + Direction rect. & \textbf{0.337} & \textbf{0.097} \\
\bottomrule
\end{tabular}
\end{table}
The direction--magnitude decomposition also allows our direction rectification to be combined with existing magnitude controllers. 
Under our decomposition, TTT3R and TTSA3R perform magnitude-only updates as in Eq.~\ref{eq:gated_update}, where the gate \(g_t^{(i)}\) is produced by each method and the candidate geometric intent \(\tilde{d}_t^{(i)}\) is left unchanged. 
We replace only this direction with the rectified direction \(\hat{d}_t^{(i)}\) of HIST3R while preserving each method's original gate:
\begin{equation}
    S_t^{(i)}
    =
    S_{t-1}^{(i)}
    +
    g_t^{(i)} \tilde{m}_t^{(i)} \hat{d}_t^{(i)} .
    \label{eq:plugin_update}
\end{equation}
As shown in Table~\ref{tab:plugin}, both hybrid variants outperform their original magnitude-only counterparts and direction rectification alone. 
This shows that direction rectification is independently effective rather than tied to a particular magnitude controller, and that it remains compatible with existing magnitude-based methods. 
The decomposition therefore not only unifies existing update-control methods, but also exposes geometric intent as a missing control axis that can be corrected independently and composed with different update-gain controllers.

\subsection{Streaming Efficiency}
\label{app:efficiency}

We evaluate streaming efficiency on TUM-Dynamics~\citep{tumd} using a single NVIDIA H100 GPU with 80GB memory. 
We vary the number of input frames from 50 to 1000 and report the peak allocated GPU memory and inference speed, measured in frames per second (FPS).

\begin{figure}[!htbp]
    \centering
    \includegraphics[width=\linewidth]{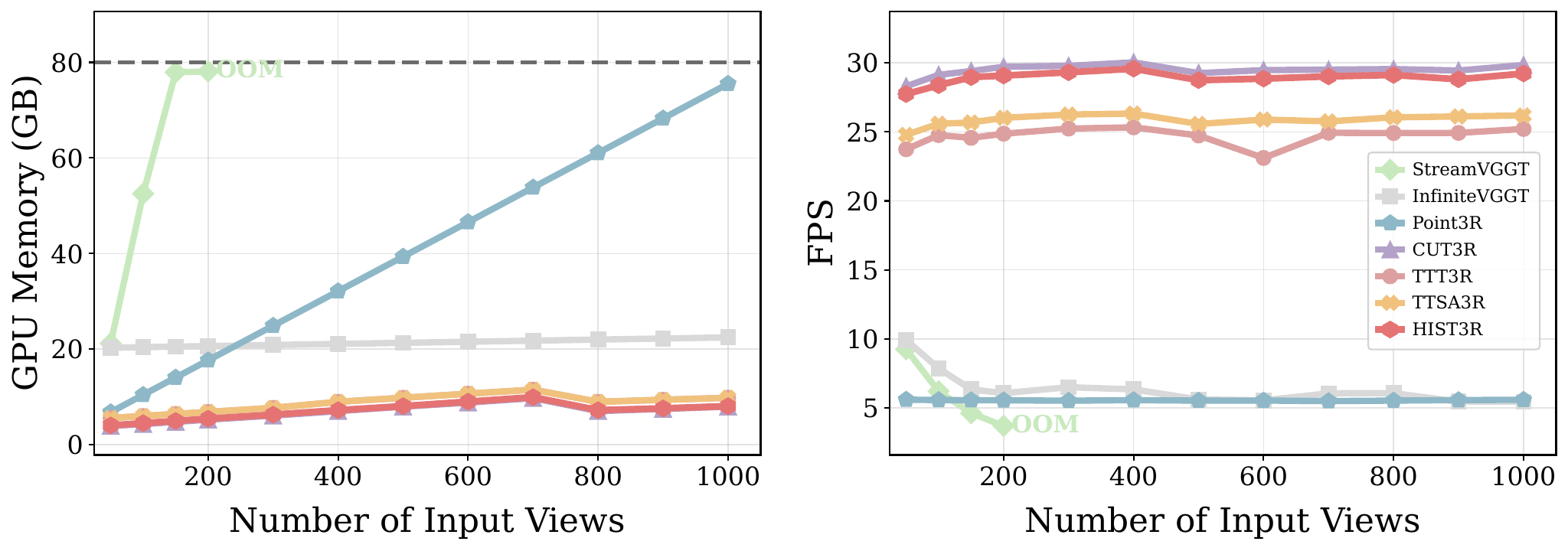}
    \caption{
        \textbf{Streaming efficiency on TUM-Dynamics~\citep{tumd}.}
        Peak allocated GPU memory (left) and inference speed measured by FPS (right) are reported under increasing input sequence length.
    }
    \label{fig:efficiency}
\end{figure}

\paragraph{Inference memory.}
As shown in Fig.~\ref{fig:efficiency}, naive explicit-memory-based methods require increasingly large memory as the sequence length grows. 
StreamVGGT quickly exceeds the 80GB memory budget, while Point3R steadily accumulates memory due to its explicit pointmap memory and approaches the memory bound at 1000 frames. 
Although InfiniteVGGT mitigates this issue through cache pruning, it still requires substantially higher memory than implicit-memory-based methods. 
In contrast, CUT3R-style implicit-memory methods, including HIST3R, maintain the lowest and nearly constant memory usage across different sequence lengths, since historical information is compressed into a compact recurrent state rather than stored as an ever-growing explicit memory.

\paragraph{Inference speed.}
In terms of inference speed, explicit-memory-based methods are substantially slower because they rely on attention over large explicit memories or caches. By maintaining historical information in a compact recurrent state, implicit-memory-based methods use much smaller attention matrices and achieve faster streaming inference.
Among CUT3R-based extensions, HIST3R maintains nearly identical FPS to the original CUT3R and is substantially faster than TTT3R and TTSA3R. 
HIST3R only exposes the state-to-image attention map from the first decoder layer for constructing visual summaries, allowing the remaining decoder layers to retain FlashAttention~\citep{flashattention}. By contrast, TTT3R and TTSA3R require attention maps from all decoder layers, which materializes attention matrices throughout inference and introduces significant overhead. 
These results show that HIST3R improves long-sequence reconstruction quality while preserving the memory and speed advantages of compact recurrent state-space inference.

\section{More Results on Additional Settings}
\label{app:additional_settings}
The main paper primarily focuses on long-sequence evaluation, where streaming methods are tested on their ability to maintain a stable scene representation over hundreds of frames. Here, we provide additional evaluations on sparse-view reconstruction and standard short-sequence benchmarks to further assess the generalization ability of HIST3R across different input regimes. The sparse-view setting tests robustness under widely separated observations with limited cross-view overlap, while the short-sequence setting enables comparison with a wider range of baselines that are infeasible for long-sequence streaming due to memory constraints. Together with the long-sequence results in the main paper, these evaluations show that the gains of HIST3R are concentrated in long-horizon streaming, while it remains competitive on standard short sequences and preserves most of the sparse-view robustness of CUT3R.

\subsection{Sparse-View Reconstruction}
\label{sec:sparse_view}
\providecommand{\oom}{\textcolor{lightgray}{\textit{OOM}}}
\definecolor{methodsepgray}{gray}{0.60}

\begin{table*}[!htbp]
\caption{
\textbf{Sparse-view 3D reconstruction results on 7-Scenes~\citep{20137scenes} and NRGBD~\citep{2022NeuralRGBD}}.
Following CUT3R~\citep{wang2025cut3r}, we sparsely sample images with stride 200 for 7-Scenes and stride 500 for NRGBD.
Best results are shown in bold, and second-best results are underlined.
}
\centering
\resizebox{\textwidth}{!}{
\begin{tabular}{llcccccccccccc}
    \toprule[0.17em]
    \multirow{3}{*}{\textbf{Method}} &
    \multirow{3}{*}{\textbf{Input}} &
    \multicolumn{6}{c}{\textbf{7-Scenes (Stride 200)}} &
    \multicolumn{6}{c}{\textbf{NRGBD (Stride 500)}} \\
    \cmidrule(lr){3-8} \cmidrule(lr){9-14}
    & &
    \multicolumn{2}{c}{\textbf{Acc. $\downarrow$}} &
    \multicolumn{2}{c}{\textbf{Comp. $\downarrow$}} &
    \multicolumn{2}{c}{\textbf{NC $\uparrow$}} &
    \multicolumn{2}{c}{\textbf{Acc. $\downarrow$}} &
    \multicolumn{2}{c}{\textbf{Comp. $\downarrow$}} &
    \multicolumn{2}{c}{\textbf{NC $\uparrow$}} \\
    \cmidrule(lr){3-4} \cmidrule(lr){5-6} \cmidrule(lr){7-8}
    \cmidrule(lr){9-10} \cmidrule(lr){11-12} \cmidrule(lr){13-14}
    & &
    \textbf{Mean} & \textbf{Med.} &
    \textbf{Mean} & \textbf{Med.} &
    \textbf{Mean} & \textbf{Med.} &
    \textbf{Mean} & \textbf{Med.} &
    \textbf{Mean} & \textbf{Med.} &
    \textbf{Mean} & \textbf{Med.} \\
    \midrule[0.08em]

    CUT3R & \multirow{4}{*}{\makecell{\textit{Sparse} \\ \textit{View}}} & \textbf{0.120} & \textbf{0.047} & \textbf{0.156} & \textbf{0.030} & \textbf{0.725} & \underline{0.831} & \textbf{0.100} & 0.036 & \textbf{0.078} & 0.031 & 0.834 & \textbf{0.970} \\
    TTT3R &  & 0.156 & 0.054 & 0.195 & 0.046 & \underline{0.719} & 0.824 & 0.105 & \underline{0.032} & \underline{0.080} & \underline{0.027} & \underline{0.835} & \textbf{0.970} \\
    TTSA3R &  & 0.152 & 0.054 & 0.189 & 0.039 & 0.718 & 0.821 & 0.107 & 0.033 & 0.082 & 0.028 & 0.834 & 
    \underline{0.968} \\
    \textbf{HIST3R} &  & \underline{0.147} & \underline{0.049} & \underline{0.176} & \underline{0.031} & \textbf{0.725} & \textbf{0.834} & \underline{0.104} & \textbf{0.031} & \underline{0.080} & \textbf{0.026} & \textbf{0.836} & \textbf{0.970} \\

    \bottomrule[0.17em]
\end{tabular}
}

\label{tab:sparse_view}
\end{table*}

Although HIST3R is designed for long-sequence streaming reconstruction, we further evaluate its robustness under sparse-view reconstruction.
Following CUT3R~\citep{wang2025cut3r}, we evaluate scene-level reconstruction using sparsely sampled images: 3 to 5 frames per scene for 7-Scenes~\citep{20137scenes} and 2 to 4 frames per scene for NRGBD~\citep{2022NeuralRGBD}.
We assess reconstruction quality by comparing predicted pointmaps with ground-truth point clouds using accuracy, completeness, and normal consistency (NC).
Accuracy measures the nearest-neighbor distance from the predicted reconstruction to the ground truth, while completeness measures the distance in the reverse direction.
For all three metrics, we report both mean and median values.

This setting is particularly challenging for history-guided updates, since sparse inputs induce large viewpoint changes with limited overlap and weaken the local temporal structure that HIST3R relies on. 
As shown in Table~\ref{tab:sparse_view}, vanilla CUT3R achieves the best overall performance, and all training-free extensions introduce some degradation. 
Among them, HIST3R consistently outperforms TTT3R and TTSA3R and remains closest to CUT3R, matching or outperforming it on 6 of the 12 metrics, including normal consistency on both datasets. 
These results indicate that HIST3R is most beneficial in dense streaming settings, while it largely preserves the robustness of CUT3R when historical structure is less reliable. 
They also show that its long-horizon gains do not come from merely conservative state updates.

\subsection{Standard Short-Sequence Evaluation}

\paragraph{Baselines.}
We include all baselines evaluated in the main paper, covering CUT3R, its training-free extensions, and explicit-memory-based methods such as Point3R, StreamVGGT, and InfiniteVGGT. To broaden the short-sequence comparison, we further include offline 3D foundation models. 
Pairwise methods, including DUSt3R~\citep{wang2024dust3r}, MASt3R~\citep{leroy2024mast3r}, MonST3R~\citep{zhang2024monst3r}, and Easi3R~\citep{chen2025easi3r}, take image pairs as input and require a global alignment stage to consolidate pairwise predictions. 
We also compare with AETHER~\citep{aether} and VGGT~\citep{wang2025vggt}, which predict all pointmaps jointly and achieve strong short-sequence reconstruction performance. 
These offline methods are effective for short sequences but are not designed for streaming inference, since adding a new frame requires rerunning inference or global alignment over the sequence. For online baselines beyond those in the main paper, we additionally include STream3R~\citep{stream3r2026} and Spann3R~\citep{wang2025spann3r}. STream3R fine-tunes DUSt3R with a causal KV-cache for streaming reconstruction, while Spann3R extends DUSt3R to online reconstruction with an external spatial memory that stores key-value features of previous frames.

\begin{table}[!htbp]
\caption{
\textbf{Evaluation on Camera Pose Estimation - Short Sequence} on Sintel~\citep{sintel}, TUM-dynamics~\citep{tumd}, and ScanNet~\citep{scannet} datasets.
HIST3R achieves the best overall performance among online methods and remains competitive with most offline baselines, despite processing frames incrementally rather than reasoning over the full sequence at once.
}
\centering
\footnotesize
\renewcommand{\arraystretch}{1.}
\renewcommand{\tabcolsep}{2.5pt}
\resizebox{\linewidth}{!}{
\begin{tabular}{@{}clcccc|ccc|ccc@{}}
\toprule
& & & \multicolumn{3}{c}{\textbf{Sintel (50 frames)}} & \multicolumn{3}{c}{\textbf{TUM-dynamics (90 frames)}} & \multicolumn{3}{c}{\textbf{ScanNet (90 frames)}} \\ 
\cmidrule(lr){4-6} \cmidrule(lr){7-9} \cmidrule(lr){10-12}
{} & {\textbf{Method}}  & \textbf{Online} & {ATE $\downarrow$} & {RPE trans $\downarrow$} & {RPE rot $\downarrow$} & {ATE $\downarrow$} & {RPE trans $\downarrow$} & {RPE rot $\downarrow$} & {ATE $\downarrow$} & {RPE trans $\downarrow$} & {RPE rot $\downarrow$} \\ 
\midrule

& Robust-CVD~\citep{Robust-CVD} &\xmark & 0.360 & 0.154 & 3.443 & 0.153 & 0.026 & 3.528 & 0.227 & 0.064 & 7.374 \\ 
 & CasualSAM~\citep{CasualSAM} &\xmark & 0.141 & \textbf{0.035} & \underline{0.615} & {0.071} & \textbf{0.010} & 1.712 & 0.158 & 0.034 & 1.618 \\ 
 & DUSt3R~\citep{wang2024dust3r} &\xmark & 0.417 & 0.250 & 5.796 & 0.083 & 0.017 & 3.567 & {0.081} & 0.028 & 0.784 \\  
{ }& MASt3R~\citep{leroy2024mast3r} &\xmark & {{0.185}} & {0.060} & {1.496} & {\underline{0.038}} & {\underline{0.012}} & {\underline{0.448}} & {{0.078}} & {{0.020}} & {\underline {0.475}} \\ 
& MonST3R~\citep{zhang2024monst3r}  &\xmark & \underline {{0.111}} & {0.044} & {0.869} & {{0.098}} & {{0.019}} & {{0.935}} & {{0.077}} & {{0.018}} & {{0.529}} \\ 

& Easi3R~\citep{chen2025easi3r}  &\xmark & \bf {{0.110}} & \underline{0.042} & {0.758} & {{0.105}} & {{0.022}} & {{1.064}} & {\underline{0.061}} & {\underline{0.017}} & {{0.525}} \\ 

& AETHER\citep{aether}  &\xmark &  {{0.189}} & {0.054} & {0.694} & {{0.092}} & {\underline{0.012}} & {{1.106}} & {{0.176}} & {{0.028}} & {{1.204}} \\ 

& VGGT~\citep{wang2025vggt}  &\xmark &  {{0.172}} & {0.062} & \bf{0.471} & \bf{{0.012}} & \bf{{0.010}} & {\bf{0.310}} & {\bf{0.035}} & {\bf{0.015}} & {\bf{0.377}} \\ 

\midrule

& Point3R~\citep{wu2025point3r} & \cmark  & 0.351 & 0.128 & 1.822 & {0.075} & 0.029 & 0.642 &{0.106} & {0.035} & {1.946}\\ 

& STream3R~\citep{stream3r2026} & \cmark  & {0.213} & 0.076 & 0.868 & \underline{0.026} & {0.013} & \underline{0.330} & \bf 0.052 &  0.021 & 0.850 \\ 

& StreamVGGT~\citep{streamvggt2026} & \cmark  & 0.251 & 0.149 & 1.894 & {0.061} & 0.033 & 3.209 &{0.161} & {0.057} & {3.647}\\ 

& InfiniteVGGT~\citep{yuan2026infinitevggt} & \cmark  & 0.227 & 0.130 & 1.669 & {0.054} & 0.031 & 2.796 & {0.146} & {0.052} & {3.290}\\ 

& Spann3R~\citep{wang2025spann3r} &\cmark   & {{0.329}} & {0.110} & {4.471} & {{0.056}} & {{0.021}} & {{0.591}} & {0.096} & {{0.023}} & {{0.661}} \\ 

& CUT3R~\citep{wang2025cut3r} & \cmark  & {0.213} & \underline{0.066} & \underline{0.621} & {0.046} & {0.015} & {0.473} & {0.099} & {0.022} & {0.600}\\ 

& TTT3R~\citep{chen2026ttt3r} & \cmark  & \underline{0.201} & \bf 0.063 & \bf 0.617 & {0.028} & \underline{0.012} & \underline{0.379} & {0.064} & {0.021} & {0.592}\\ 

& TTSA3R~\citep{zheng2026ttsa3r} & \cmark  & 0.210 & 0.085 & 0.765 & {0.028} & \underline{0.012} & \underline{0.372} & \underline{0.057} & \underline{0.020} & \underline{0.588}\\ 

& \textbf{HIST3R} & \cmark  &  \textbf{0.195} & 0.080 & 0.748 & \textbf{0.023} & \bf 0.011 & \textbf{0.329} & \underline{0.057} & \textbf{0.019} & \textbf{0.525}\\

\bottomrule
\end{tabular}
}
\label{tab:short_pose}
\end{table}

\paragraph{Camera pose estimation.}
Following prior works~\citep{zhang2024monst3r,wang2025cut3r,chen2026ttt3r}, we evaluate camera pose estimation on Sintel~\citep{sintel}, TUM-Dynamics~\citep{tumd}, and ScanNet~\citep{scannet}. 
We report Absolute Trajectory Error (ATE), Relative Translation Error (RTE), and Relative Rotation Error (RRE), after aligning the estimated camera trajectory to the ground truth with Sim(3) alignment following the standard evaluation protocol~\citep{tumd}. 
In addition to the streaming baselines considered in the main paper, we include a broader set of short-sequence baselines, including 3D reconstruction foundation models and per-sequence optimization methods such as Robust-CVD~\citep{Robust-CVD} and CasualSAM~\citep{CasualSAM}, which jointly optimize camera parameters and dense depth maps for each sequence.

The short-sequence results are reported in Table~\ref{tab:short_pose}.
Since several offline baselines rely on global optimization or full-sequence attention, they are effective in this short-sequence setting but are not directly applicable to long-sequence streaming evaluation due to memory constraints.
We therefore report offline and online methods separately.
Offline methods can access all input images simultaneously and perform global reasoning over the entire sequence, whereas online methods must process images incrementally, creating an inherent gap between the two settings.
Although HIST3R does not match the performance of the state-of-the-art offline method VGGT, it achieves the strongest overall performance among online methods.
Notably, HIST3R also surpasses several offline baselines, showing that its design remains highly competitive even under standard short-sequence evaluation.

\paragraph{Video Depth Estimation}
\begin{table}[!htbp]
\caption{
\textbf{Evaluation of video depth estimation on short sequences} on Sintel~\citep{sintel}, Bonn~\citep{bonn}, and KITTI~\citep{kitti}.
We report scale-invariant relative depth, aligned by a per-sequence scale, and metric-scale absolute depth accuracy.
HIST3R achieves the strongest metric-scale performance, while remaining competitive with leading online methods under scale-aligned evaluation on Bonn and KITTI.
}
\centering
\renewcommand{\arraystretch}{1.02}
\setlength{\tabcolsep}{1.5pt}
\resizebox{\linewidth}{!}{
\begin{tabular}{@{}llc>{\centering\arraybackslash}p{1.5cm}>{\centering\arraybackslash}p{1.5cm}|>{\centering\arraybackslash}p{1.5cm}>{\centering\arraybackslash}p{1.5cm}|>{\centering\arraybackslash}p{1.5cm}>{\centering\arraybackslash}p{1.5cm}@{}}
\toprule
 &   &  & \multicolumn{2}{c}{\textbf{Sintel (50 frames)}} & \multicolumn{2}{c}{\textbf{BONN (110 frames)}} & \multicolumn{2}{c}{\textbf{KITTI (110 frames)}} \\ 
\cmidrule(lr){4-5} \cmidrule(lr){6-7} \cmidrule(lr){8-9}
\textbf{Alignment} & \textbf{Method} & \textbf{Online } & {Abs Rel $\downarrow$} & {$\delta$\textless $1.25\uparrow$} & {Abs Rel $\downarrow$} & {$\delta$\textless $1.25\uparrow$} & {Abs Rel $\downarrow$} & {$\delta$ \textless $1.25\uparrow$} \\ 

\midrule

\multirow{11}{*}{\begin{minipage}{3cm}Per-sequence Scale\end{minipage}} 
& DUSt3R~\citep{wang2024dust3r} &\xmark & 0.656 & {45.2} & {0.155} & {83.3} & {0.144} & {81.3} \\
& MASt3R~\citep{leroy2024mast3r} &\xmark & 0.641 & {43.9} & {0.252} & {70.1} & {0.183} & {74.5} \\

& MonST3R~\citep{zhang2024monst3r} &\xmark & {0.378} & {55.8} & {0.067} & {96.3} & {0.168} & {74.4} \\

& Easi3R~\citep{chen2025easi3r} &\xmark & {0.377}  & \underline{55.9} & \underline{0.059} & \underline{97.0} & {0.102} & {91.2} \\

& AETHER~\citep{aether} &\xmark & \underline{0.324}  & {50.2} & {0.273} & {59.4} & \textbf{0.056} & \textbf{97.8} \\

& VGGT~\citep{wang2025vggt} &\xmark & \textbf{0.287}  & \textbf{66.1} & \textbf{0.055} & \textbf{97.1} & \underline{0.070} & \underline{96.5} \\

\cmidrule{2-9}

& Point3R~\citep{wu2025point3r} & \cmark & {0.452}  & {48.9} & \underline{0.060} & {96.0} & {0.136} & {84.2} \\

& STream3R~\citep{stream3r2026} & \cmark & 0.478  & \underline{51.1} & 0.075 & 94.1 & {0.116} & {89.6} \\

& StreamVGGT~\citep{streamvggt2026} & \cmark & \textbf{0.323}  & \textbf{65.7} & \textbf{0.059} & \underline{97.2} & {0.173} & {72.1} \\

& InfiniteVGGT~\citep{yuan2026infinitevggt} & \cmark & \underline{0.327}  & \textbf{65.7} & \underline{0.060} & \textbf{97.3} & {0.173} & {72.1} \\

& Spann3R~\citep{wang2025spann3r} & \cmark & 0.622 & {42.6} & {0.144} & {81.3} & {0.198} & {73.7} \\

& CUT3R~\citep{wang2025cut3r} & \cmark & {0.421}  & {47.9} & {0.078} & {93.7} & {0.118} & {88.1} \\

& TTT3R~\citep{chen2026ttt3r} & \cmark & {0.404}  & {50.0} & {0.068} & {95.4} & \underline{0.113} & {90.4} \\

& TTSA3R~\citep{zheng2026ttsa3r} & \cmark & {0.401}  & {50.0} & {0.064} & {96.5} & \textbf{0.110} & \textbf{91.2} \\

& \textbf{HIST3R} & \cmark & {0.408}  & {50.1} & {0.061} & {96.9} & \textbf{0.110} & \underline{90.8} \\

\midrule
 \multirow{4}{*}{\begin{minipage}{3cm}Metric Scale\end{minipage}} 
& MASt3R~\citep{leroy2024mast3r} &\xmark & {1.022}  & 14.3 & 0.272 & 70.6 & 0.467 & 15.2 \\  
 
& Point3R~\citep{wu2025point3r} & \cmark & \textbf{{0.777}} & {17.1} & {0.137} & {94.7}& {0.191} & 73.8 \\

& STream3R~\citep{stream3r2026} & \cmark & 1.041 & 21.0 & {0.084} & {94.4} & 0.234 & 57.6 \\
 
& CUT3R~\citep{wang2025cut3r} & \cmark & {1.029} & {23.8} & {0.103} & {88.5}& {0.122} & {85.5} \\

& TTT3R~\citep{chen2026ttt3r} & \cmark & {0.977} & {24.5} & {0.090} & 94.2 & \textbf{0.110} & \underline{89.1}  \\

& TTSA3R~\citep{zheng2026ttsa3r} & \cmark & {0.960} & \underline{24.6} & \underline{0.079} & \underline{96.6} & \underline{0.111} & {88.9}  \\

& \textbf{HIST3R} & \cmark & \underline{0.935} & \textbf{25.3} & \textbf{0.068} & \textbf{96.7} & \textbf{0.110} & \textbf{90.8}  \\

\bottomrule
\end{tabular}
}

\label{tab:short_depth}
\end{table}

Following common practice~\citep{zhang2024monst3r, wang2025cut3r, chen2026ttt3r}, we evaluate video depth estimation on KITTI~\citep{kitti}, Sintel~\citep{sintel}, and Bonn~\citep{bonn} datasets covering dynamic and static, indoor and outdoor scenes. 
We report absolute relative error (Abs Rel) and $\delta < 1.25$, which measures the percentage of pixels whose predicted depth is within a factor of $1.25$ of the ground truth.
To evaluate relative depth quality and temporal consistency, we align predicted depths to the ground truth with a single scale factor per sequence.
For methods that output metric pointmaps, we additionally report results without scale alignment, directly evaluating the predictions in metric units to measure absolute-scale accuracy.

As shown in Table~\ref{tab:short_depth}, HIST3R achieves the strongest overall metric-scale performance.
Under scale-aligned evaluation, HIST3R remains competitive with leading online methods such as StreamVGGT and InfiniteVGGT on Bonn and KITTI, while being substantially more efficient in both memory usage and inference speed.
Together with the long-sequence and sparse-view results, these results show that the benefits of HIST3R are not limited to long-horizon streaming reconstruction; its update design also remains effective under standard short-sequence evaluation and challenging sparse-input scenarios.

\section{Qualitative Results}

We provide qualitative results for the long-horizon streaming reconstruction setting studied in the main paper.
Specifically, we visualize both 3D reconstruction and camera pose estimation results under 500-frame inputs, which represent a challenging sequence length for existing online reconstruction methods.

\begin{figure}[!p]
\centering
\makebox[\linewidth][c]{%
\includegraphics[height=0.44\textheight]{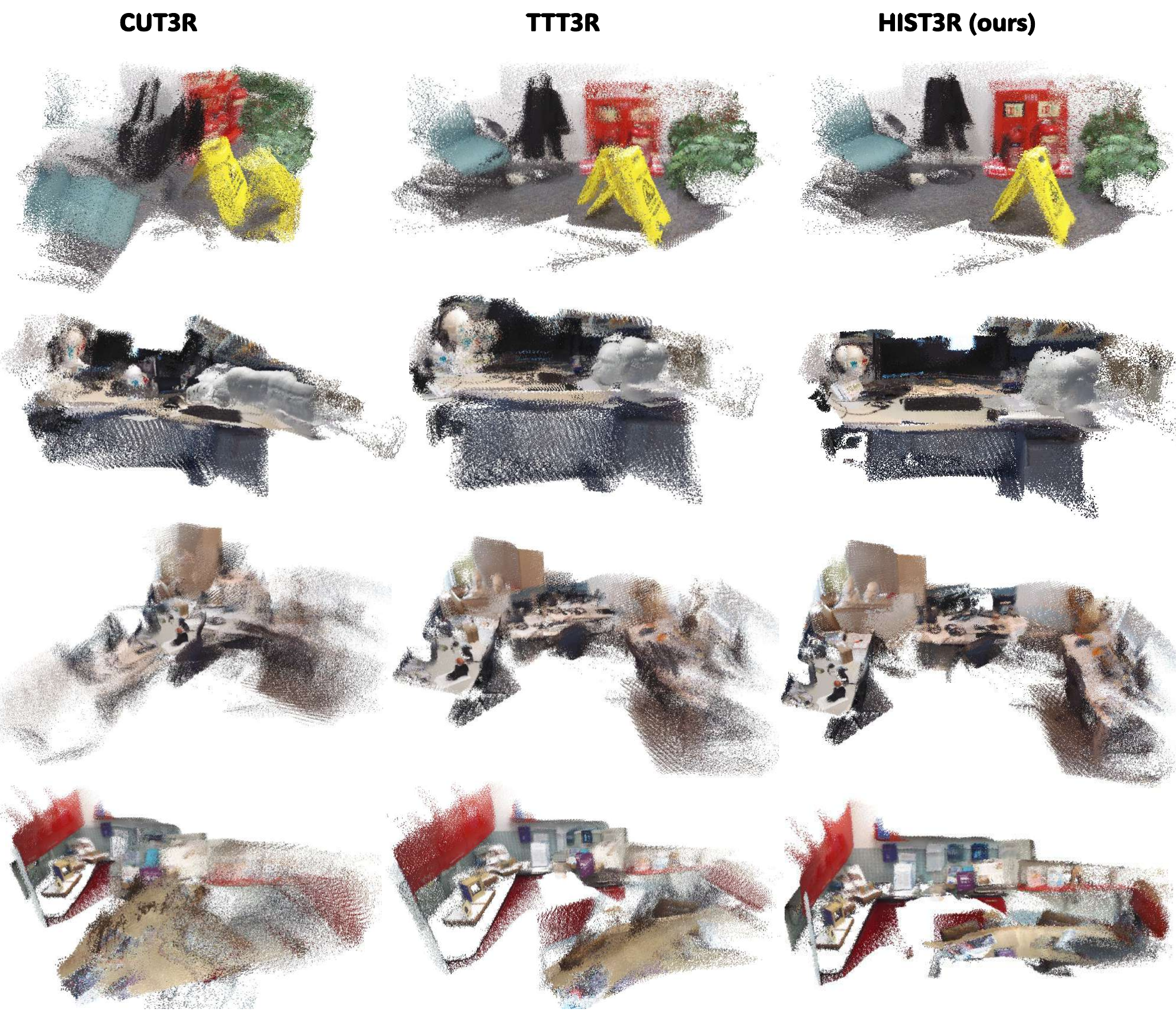}
}
\caption{
\textbf{Qualitative comparison with implicit-memory-based methods on 7-Scenes.}
CUT3R exhibits clear long-horizon degradation, while HIST3R substantially improves reconstruction coherence and reduces accumulated artifacts.
}

\label{fig:qual_7scenes_imp}
\vspace{-8pt}
\end{figure}

\begin{figure}[!p]
\centering
\makebox[\linewidth][c]{%
\includegraphics[height=0.405\textheight]{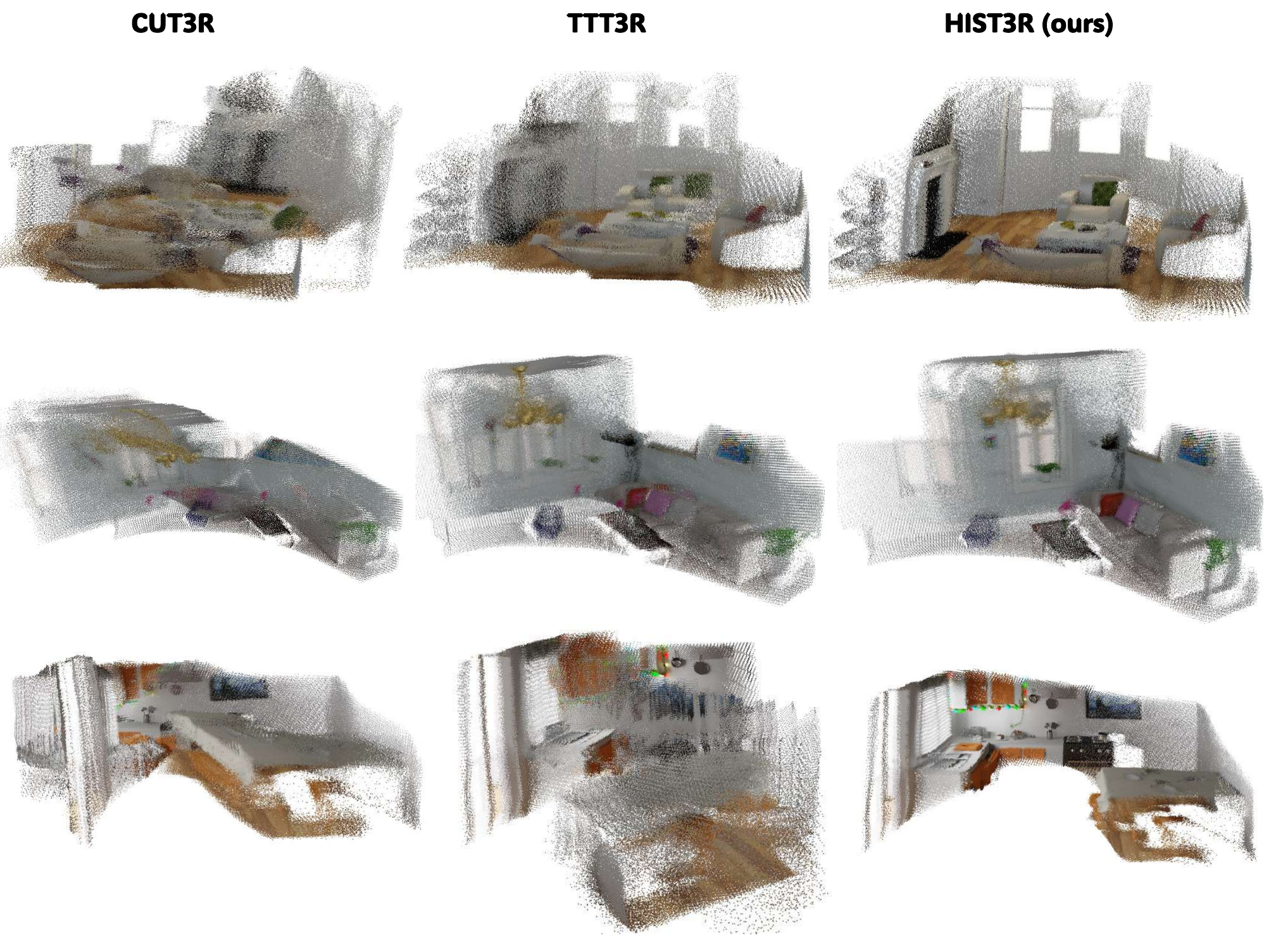}
}
\caption{
\textbf{Qualitative comparison with implicit-memory-based methods on NRGBD.}
TTT3R partially improves over CUT3R, but still exhibits severe accumulated artifacts and collapsed scene geometry, whereas HIST3R produces substantially more coherent and stable reconstructions.
}
\label{fig:qual_nrgbd_imp}
\vspace{-8pt}
\end{figure}

\begin{figure}[!p]
\centering
\makebox[\linewidth][c]{%
\includegraphics[height=0.44\textheight]{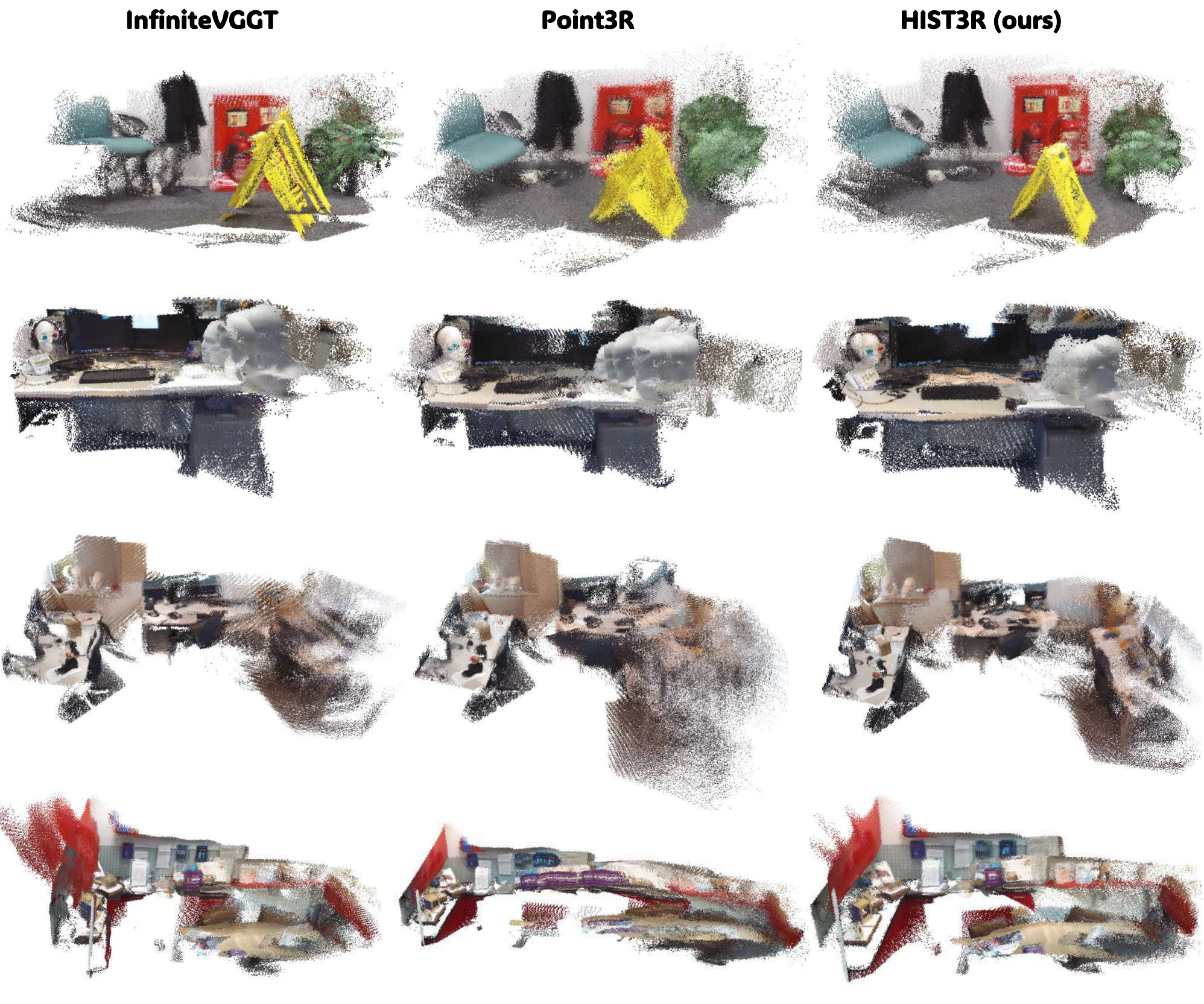}
}
\caption{
\textbf{Qualitative comparison with explicit-memory-based methods on 7-Scenes.}
Point3R exhibits noticeable geometric distortion, while InfiniteVGGT remains competitive with HIST3R but still produces slightly blurred reconstructions in some regions.
}
\label{fig:qual_7scenes_exp}
\vspace{-8pt}
\end{figure}

\begin{figure}[!p]
\centering
\makebox[\linewidth][c]{%
\includegraphics[height=0.39\textheight]{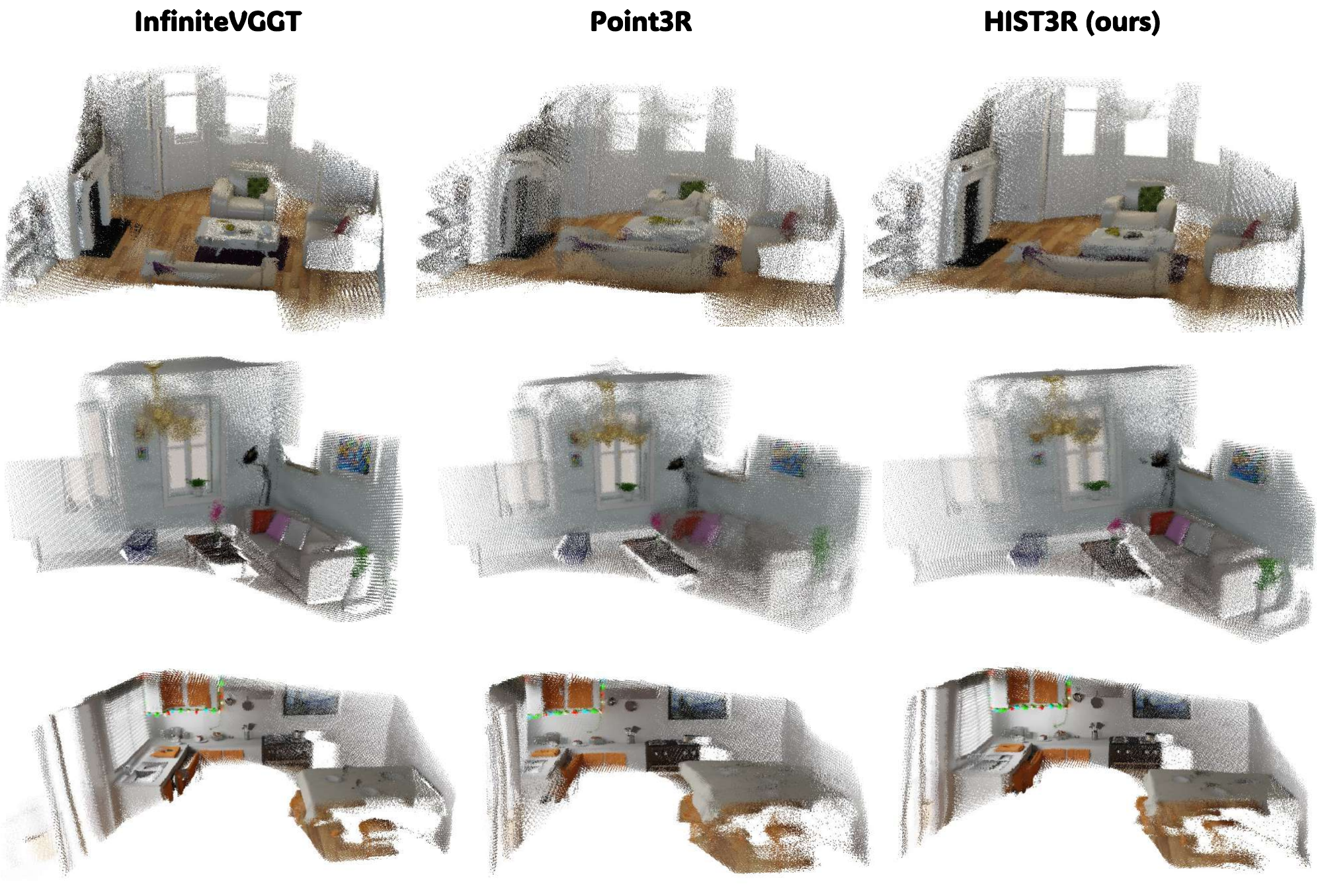}
}
\caption{
\textbf{Qualitative comparison with explicit-memory-based methods on NRGBD.}
HIST3R and InfiniteVGGT produce more stable reconstruction quality than Point3R, with cleaner local geometry and fewer visible distortions.
}
\label{fig:qual_nrgbd_exp}
\vspace{-8pt}
\end{figure}

\paragraph{3D Reconstruction.}
We provide qualitative comparisons on 7-Scenes and NRGBD under the 500-frame streaming setting.
To make the comparison clearer, we organize the results according to the type of online memory used by the baselines: implicit-memory methods and explicit-memory methods. For implicit-memory methods, Fig.~\ref{fig:qual_7scenes_imp} and Fig.~\ref{fig:qual_nrgbd_imp} show that CUT3R suffers from severe long-horizon degradation, leading to blurred and inconsistent reconstructions.
TTT3R partially alleviates this degradation, but still exhibits visible artifacts and structural drift.
In contrast, HIST3R produces cleaner and more stable reconstructions, demonstrating the benefit of direction--magnitude update control for maintaining scene geometry over long sequences.
We further compare with explicit-memory-based methods in Fig.~\ref{fig:qual_7scenes_exp} and Fig.~\ref{fig:qual_nrgbd_exp}. Point3R can produce distorted scene geometry in some cases.
HIST3R remains competitive with InfiniteVGGT while preserving stable scene structure, suggesting its ability to maintain strong long-horizon reconstruction quality without relying on explicit memory storage.

\begin{figure}[!t]
    \centering
    \includegraphics[width=1\linewidth]{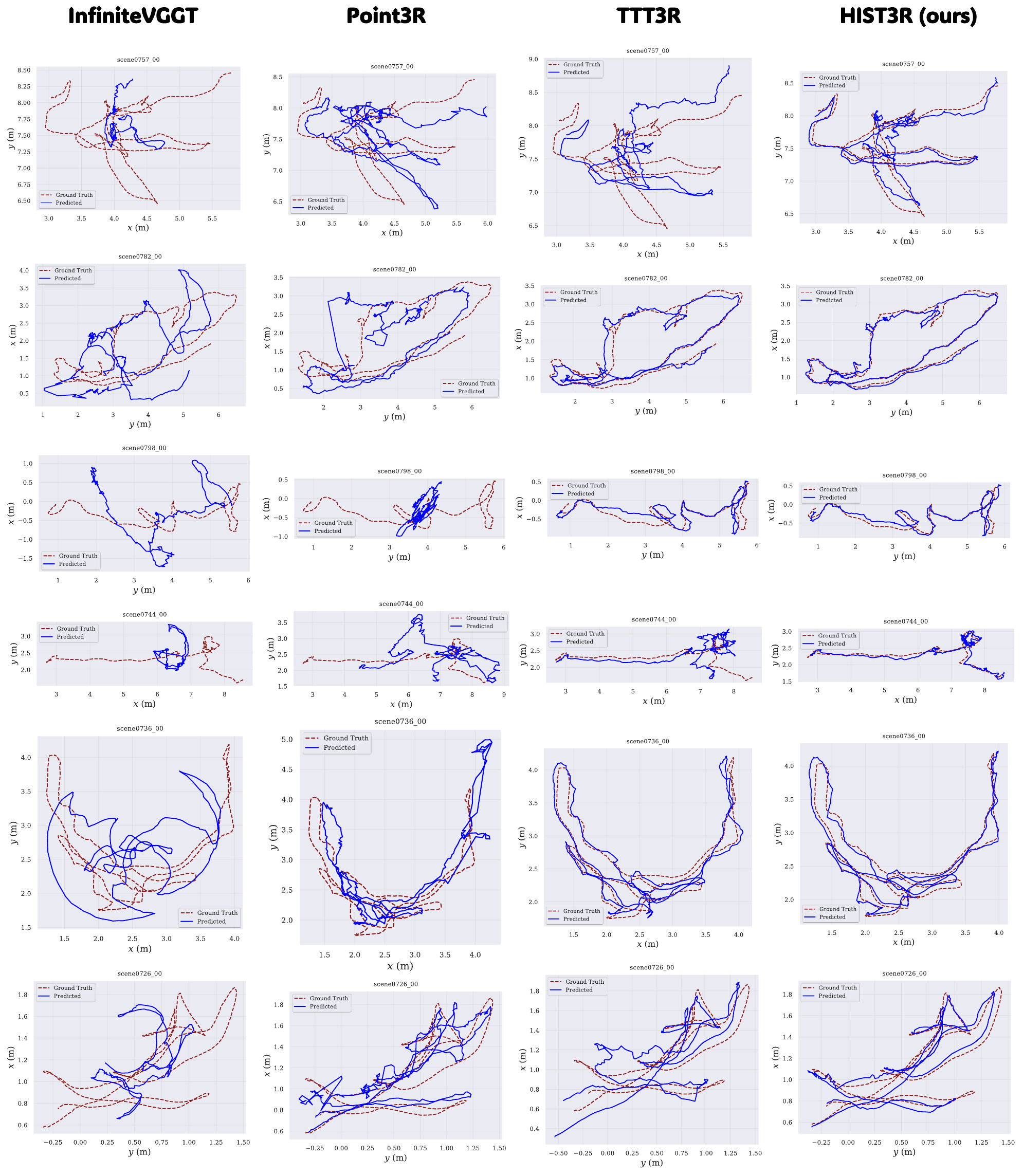}
    \caption{
        \textbf{Qualitative trajectory results on ScanNet.}
        Blue curves denote predicted trajectories, and red dashed curves denote ground-truth trajectories.
        Explicit-memory-based methods exhibit noticeable pose drift over long sequences.
        TTT3R produces more stable trajectories, but still shows a clear gap compared with HIST3R, which better preserves long-horizon trajectory consistency.
    }
    \label{fig:qual_scannet}
\end{figure}

\begin{figure}[!htbp]
    \centering
    \includegraphics[width=0.95\linewidth]{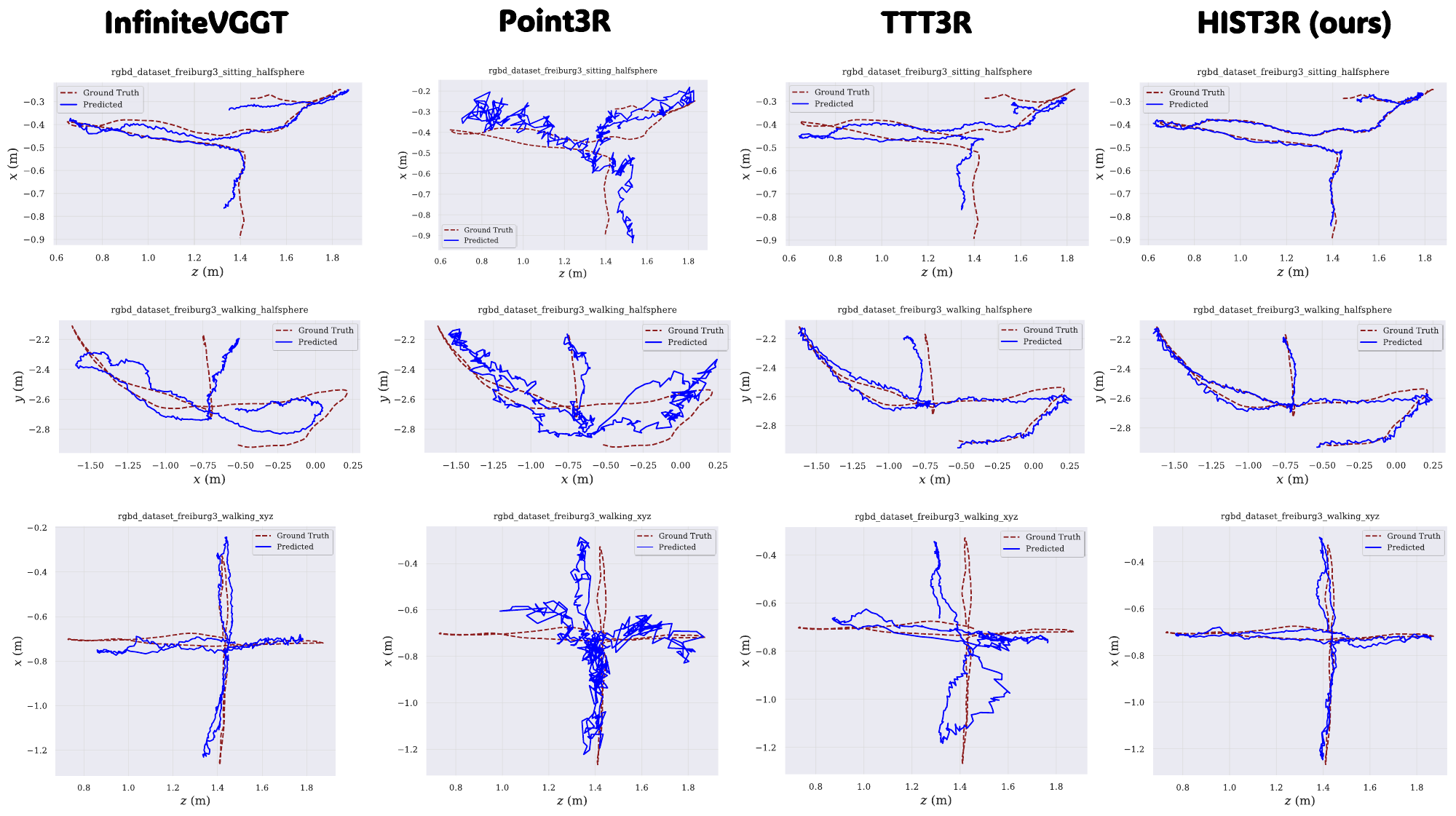}
    \caption{
        \textbf{Qualitative trajectory results on TUM-Dynamics.}
        HIST3R shows a clear advantage over all compared methods, preserving substantially better long-horizon trajectory consistency.
    }
    \label{fig:qual_tum}
\end{figure}

\paragraph{Camera Pose Estimation.}
We provide qualitative trajectory comparisons on ScanNet and TUM-Dynamics in Fig.~\ref{fig:qual_scannet} and Fig.~\ref{fig:qual_tum}.
Existing online baselines often accumulate visible trajectory drift over long sequences.
HIST3R, in contrast, closely follows the ground-truth trajectory, demonstrating improved long-horizon pose stability under streaming inference.

\clearpage

\section{Discussion}
\label{sec:discussion}

\paragraph{Limitations.}
As a training-free method, HIST3R is ultimately bounded by the representation and candidate updates produced by the pretrained CUT3R model.
Although HIST3R improves recurrent state updates by modulating the direction and magnitude of each candidate update at inference time, it cannot recover geometric information that is not encoded by the underlying model or fully correct failures caused by severely inaccurate candidate predictions.
A promising direction is to incorporate this direction--magnitude view into pretraining or fine-tuning, so that the model can learn to produce more reliable geometric intent and update gain rather than correcting them only at inference time.

\paragraph{Broader impact.}
This work analyzes the evolution of implicit recurrent states in streaming 3D reconstruction from a geometric perspective.
By decomposing state updates into direction and magnitude, HIST3R provides an interpretable view of how online reconstruction models accumulate, preserve, and revise geometric information over long sequences.
We believe this perspective can inform the design of future state-space reconstruction models, encouraging update mechanisms that explicitly reason about both geometric intent and update strength.
Such models may benefit online 3D perception systems that require stable long-horizon reconstruction under bounded memory, including embodied AI, augmented reality, and robotics.
At the same time, advances in streaming reconstruction may strengthen downstream applications involving spatial understanding and scene capture, and should therefore be deployed with appropriate consideration of privacy and safety when reconstructing real-world environments.

\newpage
\section*{NeurIPS Paper Checklist}

\begin{enumerate}

\item {\bf Claims}
    \item[] Question: Do the main claims made in the abstract and introduction accurately reflect the paper's contributions and scope?
\item[] Answer: \answerYes{}.
\item[] Justification: The abstract and introduction accurately describe HIST3R as a training-free extension of CUT3R that decomposes recurrent updates into direction and magnitude. The stated claims are supported by experiments across multiple 3D perception tasks, and the scope is clearly limited to inference-time improvement over pretrained streaming reconstruction models.

\item {\bf Limitations}
    \item[] Question: Does the paper discuss the limitations of the work performed by the authors?
    \item[] Answer: \answerYes{}.
    \item[] Justification: We discuss limitations in Appendix~\ref{sec:discussion}, noting that HIST3R is a training-free method whose performance is bounded by the pretrained CUT3R model, and that future work could incorporate the direction--magnitude decomposition into pretraining or fine-tuning. We also discuss in Appendix~\ref{sec:sparse_view} that HIST3R is most beneficial in dense streaming settings and slightly underperforms CUT3R under sparse-view inputs, where the local temporal structure it relies on is weaker.

\item {\bf Theory assumptions and proofs}
    \item[] Question: For each theoretical result, does the paper provide the full set of assumptions and a complete (and correct) proof?
    \item[] Answer: \answerNA{}.
    \item[] Justification: The paper does not present formal theoretical results or proofs. The self-expressive and direction--magnitude decomposition views are used as methodological motivation and are supported empirically through ablations and analysis.

    \item {\bf Experimental result reproducibility}
    \item[] Question: Does the paper fully disclose all the information needed to reproduce the main experimental results of the paper to the extent that it affects the main claims and/or conclusions of the paper (regardless of whether the code and data are provided or not)?
    \item[] Answer: \answerYes{}.
    \item[] Justification: We describe the evaluation protocols, datasets, baselines, and metrics in Sec.~\ref{sec:exp_pose_estimation}--\ref{sec:ablation}, and the full method in Sec.~3. The selection protocol and final values of all hyperparameters, including the history window size and the alignment and discrepancy gate parameters, are reported in Appendix~\ref{app:hyperparameters}, together with a sensitivity analysis. Design ablations are provided in Sec.~\ref{sec:ablation}.

\item {\bf Open access to data and code}
    \item[] Question: Does the paper provide open access to the data and code, with sufficient instructions to faithfully reproduce the main experimental results, as described in supplemental material?
\item[] Answer: \answerNo{}.
\item[] Justification: The code is not publicly released at the time of publication. All experiments use publicly available datasets, and we describe the evaluation protocols, metrics, baselines, and hyperparameters needed to reproduce the main results.

\item {\bf Experimental setting/details}
    \item[] Question: Does the paper specify all the training and test details (e.g., data splits, hyperparameters, how they were chosen, type of optimizer) necessary to understand the results?
\item[] Answer: \answerYes{}.
\item[] Justification: The paper specifies the test protocols, datasets, metrics, and baselines in Sec.~\ref{sec:exp_pose_estimation}--\ref{sec:ablation}. All hyperparameters are selected once on 30 randomly sampled sequences from the ScanNet training split without using test data, and kept fixed across all experiments; their values and selection protocol are reported in Appendix~\ref{app:hyperparameters}. Since HIST3R is training-free, no optimizer or training schedule is required.

\item {\bf Experiment statistical significance}
    \item[] Question: Does the paper report error bars suitably and correctly defined or other appropriate information about the statistical significance of the experiments?
\item[] Answer: \answerNo{}.
\item[] Justification: We do not report error bars or statistical significance tests. Our evaluation follows prior streaming reconstruction work and reports deterministic benchmark metrics across multiple datasets, sequence lengths, and tasks.

\item {\bf Experiments compute resources}
    \item[] Question: For each experiment, does the paper provide sufficient information on the computer resources (type of compute workers, memory, time of execution) needed to reproduce the experiments?
\item[] Answer: \answerYes{}.
\item[] Justification: The GPU type and memory budget are reported in Sec.~\ref{sec:experiments} and Appendix~\ref{app:efficiency}, where all experiments are conducted on a single NVIDIA H100 GPU with 80GB memory. Since HIST3R is training-free, no training compute is required; inference-time memory and speed are reported in Appendix~\ref{app:efficiency} and Fig.~\ref{fig:efficiency}.

\item {\bf Code of ethics}
    \item[] Question: Does the research conducted in the paper conform, in every respect, with the NeurIPS Code of Ethics \url{https://neurips.cc/public/EthicsGuidelines}?
\item[] Answer: \answerYes{}.
\item[] Justification: The research uses standard public 3D reconstruction benchmarks and does not involve human-subject studies, private data collection, or deployment on sensitive real-world systems. We also discuss broader impact and potential privacy considerations for real-world reconstruction applications.

\item {\bf Broader impacts}
    \item[] Question: Does the paper discuss both potential positive societal impacts and negative societal impacts of the work performed?
\item[] Answer: \answerYes{}.
\item[] Justification: The paper discusses broader impacts in the appendix, including potential benefits for online 3D perception systems such as robotics, embodied AI, and AR/VR, as well as privacy and safety considerations when reconstructing real-world environments.
    
\item {\bf Safeguards}
    \item[] Question: Does the paper describe safeguards that have been put in place for responsible release of data or models that have a high risk for misuse (e.g., pre-trained language models, image generators, or scraped datasets)?
\item[] Answer: \answerNA{}.
\item[] Justification: The paper does not release new high-risk datasets or pretrained generative models. The method is evaluated on standard public 3D reconstruction benchmarks and builds on existing reconstruction models.

\item {\bf Licenses for existing assets}
    \item[] Question: Are the creators or original owners of assets (e.g., code, data, models), used in the paper, properly credited and are the license and terms of use explicitly mentioned and properly respected?
\item[] Answer: \answerYes{}.
\item[] Justification: The paper cites the original works for all datasets, baselines, and pretrained models used in the experiments. We use standard public benchmarks and existing models according to their intended research use.

\item {\bf New assets}
    \item[] Question: Are new assets introduced in the paper well documented and is the documentation provided alongside the assets?
\item[] Answer: \answerNA{}.
\item[] Justification: The paper does not introduce new datasets or benchmark assets. HIST3R is a training-free method evaluated on existing public datasets and baselines.

\item {\bf Crowdsourcing and research with human subjects}
    \item[] Question: For crowdsourcing experiments and research with human subjects, does the paper include the full text of instructions given to participants and screenshots, if applicable, as well as details about compensation (if any)? 
\item[] Answer: \answerNA{}.
\item[] Justification: The paper does not involve crowdsourcing experiments or research with human subjects.

\item {\bf Institutional review board (IRB) approvals or equivalent for research with human subjects}
    \item[] Question: Does the paper describe potential risks incurred by study participants, whether such risks were disclosed to the subjects, and whether Institutional Review Board (IRB) approvals (or an equivalent approval/review based on the requirements of your country or institution) were obtained?
\item[] Answer: \answerNA{}.
\item[] Justification: The paper does not involve research with human subjects, crowdsourcing, or participant studies; therefore, IRB approval is not applicable.

\item {\bf Declaration of LLM usage}
    \item[] Question: Does the paper describe the usage of LLMs if it is an important, original, or non-standard component of the core methods in this research? Note that if the LLM is used only for writing, editing, or formatting purposes and does \emph{not} impact the core methodology, scientific rigor, or originality of the research, declaration is not required.
\item[] Answer: \answerNA{}.
\item[] Justification: LLMs are not used as an important, original, or non-standard component of the core method. Any LLM usage, if applicable, is limited to writing, editing, or formatting assistance and does not affect the methodology, experiments, or scientific claims.

\end{enumerate}

\end{document}